\documentclass[preprint, review,3p,times, 10pt]{elsarticle}

\usepackage{amsmath,amssymb,amsfonts}
\usepackage{graphicx}
\usepackage{textcomp}
\usepackage{comment}
\usepackage{footnote}
\usepackage{threeparttable}
\usepackage{romannum}
\usepackage{multirow}
\usepackage{longtable}
\usepackage{lscape}
\usepackage{soul}
\usepackage{array, makecell}
\usepackage{pifont}
\usepackage{booktabs}
\usepackage{pgfplots}
\usepackage{tabularx}
\usepackage{tikz}
\usepackage{listings}
\usepackage[utf8]{inputenc}
\usepackage[T1]{fontenc}
\DeclareUnicodeCharacter{2032}{\ensuremath{\prime}}
\usepackage{pgf-pie}
\usepackage[ruled,vlined]{algorithm2e}
\usepackage{float}
\usepackage[hyphens]{url}
\usepackage{hyperref}
\hypersetup{colorlinks=true,breaklinks=true,hypertexnames=false}
\usepackage{enumitem}
\usepackage{caption}
\usepackage{subcaption}
\pgfplotsset{compat=1.18}

\journal{''}

\begin{document}

\begin{frontmatter}
\title{BenSParX: A Robust Explainable Machine Learning Framework for Parkinson's Disease Detection from  Bengali Conversational Speech}
\author[1,2]{Riad Hossain}

\author[3]{Muhammad Ashad Kabir \corref{correspondingauthor}}
\cortext[correspondingauthor]{Corresponding author: Charles Sturt University, Panorama Ave, Bathurst, NSW 2795. Ph.+61263386259}
\ead{akabir@csu.edu.au}
\author[1]{Arat Ibne Golam Mowla}
\author[1]{Animesh Chandra Roy}
\author[4]{Ranjit Kumar Ghosh}
\affiliation[1]{organization={Department of Computer Science and Engineering, Chittagong University of Engineering and Technology}, city={Chittagong}, postcode={4349},country={Bangladesh}}
 \affiliation[2]{organization={School of Science, Engineering and Technology, East Delta University}, city={Chittagong}, postcode={4209},  country={Bangladesh}}
 \affiliation[3]{organization={School of Computing, Mathematics and Engineering, Charles Sturt University}, city={Bathurst}, state={NSW}, postcode={2795}, country={Australia}}
\affiliation[4]{organization={Department of Neurology,
Bangabandhu Sheikh Mujib Medical University}, city={}, country={Bangladesh}}
\begin{abstract}

Parkinson's disease (PD) poses a growing global health challenge, with Bangladesh experiencing a notable rise in PD-related mortality. Early detection of PD remains particularly challenging in resource-constrained settings, where voice-based analysis has emerged as a promising non-invasive and cost-effective alternative. However, existing studies predominantly focus on English or other major languages; notably, no voice dataset for PD exists for Bengali -- a language spoken by over 230 million people worldwide -- posing a significant barrier to culturally inclusive and accessible healthcare solutions.
Moreover, most prior studies employed only a narrow set of acoustic features, with limited or no hyperparameter tuning and feature selection, and little attention to model explainability. This restricts the development of a robust and generalizable machine learning (ML) model.
To address this gap, we present \textit{BenSparX}, the first Bengali conversational speech dataset for PD detection, along with a robust and explainable ML framework tailored for early diagnosis. The proposed framework incorporates diverse acoustic feature categories, systematic feature selection methods, and state-of-the-art ML classifiers with extensive hyperparameter optimization. Furthermore, to enhance interpretability and trust in model predictions, the framework incorporates SHAP (SHapley Additive exPlanations) analysis to quantify the contribution of individual acoustic features toward PD detection. Our framework achieves state-of-the-art performance, yielding an accuracy of 95.67\%, F1 score of 95.62\%, and AUC of 0.990. We further validated our approach by applying the framework to existing PD datasets in other languages, where it consistently outperforms state-of-the-art approaches.
This study lays the foundation for identifying subtle yet clinically meaningful vocal biomarkers, particularly in low-resource settings such as Bengali-speaking populations, and represents a significant step toward equitable, explainable, and robust digital health diagnostics for neurodegenerative disorders. The labelled dataset created in this study is available at \url{https://github.com/riadEDU/BenSParX}.
\end{abstract}

\end{frontmatter}

\section{Introduction}
Parkinson's disease (PD) is the second most prevalent neurological disorder, affecting over 10 million older adults worldwide~\citep{9426437,prashanth2018early,standaert2015parkinson}, and results from the degeneration of dopaminergic neurons in the midbrain~\citep{parkinson2002essay}. In Bangladesh, PD-related mortality has risen 46.6\% since 1990, with an average annual growth of 2\%, predominantly affecting men over 80, though women aged 75–79 exhibit the highest mortality rate~\citep{CiteDrive2022}. According to WHO, 3,782 deaths (0.53\% of all deaths) were attributed to PD in 2020~\citep{CiteDrive2024}.

PD patients commonly show bradykinesia, rigidity, tremors, impaired balance, and dysphonia~{
\citep{berardelli2001pathophysiology, jankovic2005motor,sakar2013collection,shahed2007motor}}
, with vocal deficits among the earliest signs~{
\citep{ho1999speech}}
. Speech impairments, including breathy voice, low volume, unclear articulation, prolonged pauses, and monotone delivery, occur in 70–90\% of cases~\citep{MA20201,NGO2022107133,lim2025cross,favaro2023interpretable}. Acoustic analysis enables early detection of subtle voice abnormalities, often imperceptible to human listeners~{
\citep{ali2019automated,harel2004acoustic,iyer2023machine, postuma2016voice}}
. Most research focuses on English and a few major languages, overlooking linguistic diversity and limiting global applicability~\citep{pinto2024acoustic}.
Numerous studies have explored PD detection through voice analysis in various languages, including Spanish~\citep{9426437}, English~\citep{di2024machine, 9887934, swain2024towards, tsanas2012accurate}, Mandarin~\citep{wang2022early}, Czech~\citep{scimeca2023robust}, Portuguese~\citep{braga2019automatic, proencca2014characterizing}, Turkish~\citep{el2022enhancing, nissar2019voice, 10.1145/3441417.3441425}, and Italian~\citep{9650872, scimeca2023robust}. However, to date, no studies have addressed PD detection using voice data in the Bengali language. This is a critical gap, as speech characteristics vary significantly across languages, and models trained in one language often fail to generalize to others.
For instance, Mandarin speakers typically pause more frequently while speaking -- not due to motor symptoms but due to the language's cognitive demands -- which can be misinterpreted as PD-related dysfluency~\citep{wang2022early}. Similarly, differences in breath control and speech rhythm between Chinese and English speakers can lead to misclassification~\citep{wang2022early}.

Additionally, most prior research has focused on a narrow set of acoustic features. Mel-Frequency Cepstral Coefficients (MFCCs) are among the most widely used in PD detection~\citep{9650872, braga2019automatic, di2024machine, 9887934, tsanas2012accurate, wang2022early}. Some studies have also included jitter, shimmer, and harmonicity~\citep{9887934, NARANJO2017147, 7042180, parisi2021m, ecsa-11-20481, swain2024towards, tsanas2012accurate}, while a smaller subset has considered pitch and intensity variations~\citep{9426437, sakar2013collection, tsanas2012accurate}. However, few, if any, studies have conducted a comprehensive analysis of formants, pulse-related features, and a diverse combination of acoustic parameters, particularly in the context of Bengali speech.
Furthermore, existing studies employed limited or no feature selection strategies and hyperparameter tuning in the model development, and paid little attention to the model explainability. Moreover, many existing studies lack rigorous feature selection strategies and often rely on default or minimal hyperparameter tuning during model development. In addition, model explainability is frequently overlooked, which limits the transparency and clinical interpretability of the diagnostic outcomes.

To address these gaps, this study introduces BenSParX, a novel Bengali conversational speech dataset and a linguistically inclusive machine learning (ML) framework for PD detection. The framework integrates a comprehensive set of acoustic features, including formants, spectrogram, pulse-related metrics, pitch, intensity, jitter, shimmer, and harmonicity. This diverse feature set captures both prosodic and fine-grained articulatory characteristics, enabling a more accurate representation of PD-related speech impairments across linguistic and demographic contexts.

A key strength of our framework is its systematic multi-stage feature selection pipeline, which combines RFECV, LASSO, Relief-F with Sequential Forward Selection (SFS), and the Mann-Whitney U test. This approach ensures the selection of the most informative and non-redundant features, reducing overfitting and improving generalizability. Additionally, extensive hyperparameter tuning is performed across multiple ML models using random search and cross-validation to identify optimal configurations.
Our proposed framework is rigorously evaluated against state-of-the-art PD detection methods, including validation on non-Bengali datasets, demonstrating improved accuracy, efficiency, and cross-linguistic generalizability.
To enhance transparency and clinical applicability, we incorporate SHapley Additive exPlanations (SHAP) for model interpretability. This enables quantification of individual feature contributions, providing insights into the most influential acoustic markers for PD detection. By combining strong predictive performance with explainability, our framework supports reliable and interpretable voice-based screening, particularly for underrepresented linguistic populations.

\section{Related work}
\subsection{Parkinson's voice datasets}
A wide range of voice-based datasets has been developed to support machine learning research in the automatic detection of PD. These datasets differ in several key aspects, including language, sample size, modality (e.g., sustained vowels, words, sentences), and recording environment (e.g., clinical, real-world, or noise-controlled settings). Table~\ref{tab:summary of dataset} provides a comprehensive summary of widely used PD voice datasets.
\begin{table}[!ht]
\centering
\caption{Summary of Parkinson's voice datasets. \label{tab:summary of dataset}}
\resizebox{1\textwidth}{!}{%
\begin{tabular}{llcccll}
\toprule
\multirow{2}{*}{Reference} & \multirow{2}{*}{Participant language} & \multicolumn{2}{c}{Subject count}& \multirow{2}{*}{Availability} & \multirow{2}{*}{Modality} &\multirow{2}{*}{Recording environment}\\ \cmidrule(r){3-4}
& & PD & HC & & &\\ \hline
\midrule
\citet{9887934} & Australian & 36 & 36 &\ding{55} & Phonemes /a/, /o/, /m/  & Clinical setting\\

\citet{viswanathan2019complexity} & Australian & 24 & 22 &\ding{55} & Phonemes /a/, /u/, /m/ & Clinical setting\\

ItalianPVS \citep{dimauro2017assessment} & Italian & 28 & 37 &\ding{55} & Balanced text and balanced words & Echo free room \\

\citet{9650872} & Italian  & 54 & 40 &\ding{55} &Vowels and balanced words & Real-world condition\\

\citet{scimeca2023robust} & Italian, Spanish, Czech & 178 & 176 & \ding{55}& Vowels /a/, /e/ and phrases  & Clinical setting\\

UCI \citep{early_biomarkers_of_parkinsons_disease_based_on_natural_connected_speech_392} & Czech & 30 & 50 &\ding{51}\textsuperscript{\href{https://archive.ics.uci.edu/dataset/392/early+biomarkers+of+parkinson+s+disease+based+on+natural+connected+speech}{a} }& Balanced text & Quiet room\\

\citet{jeong2024machine} &Korean  & 100 & 100 &\ding{55} &Vowels and Korean consonants & Noise controlled \\

\citet{wang2022early} & Mandarin & 50 & 50 & \ding{55}&Sustained vowels\textsuperscript{b} and sentences
& Real-world condition\\

CPPDD \citep{ZhangTao59} & Mandarin & 40 & 40 & \ding{55}&Chinese vowels 'a', 'o'
& Controlled environment\\

 \citet{proencca2014characterizing} &  Portuguese & 22 & 30 &\ding{55}& Sentences and words & Clinical setting\\

Sakar18
 \citep{parkinsons_disease_classification_470} & Turkish & 188 & 64 & \ding{51}\textsuperscript{\href{https://archive.ics.uci.edu/dataset/470/parkinson+s+disease+classification}{c}} & Vowel /a/  & Clinical setting\\

 Sakar13 \citep{parkinsons_speech_with_multiple_types_of_sound_recordings_301} & Turkish & 20 & 20 &\ding{51}\textsuperscript{\href{https://archive.ics.uci.edu/dataset/301/parkinson+speech+dataset+with+multiple+types+of+sound+recordings}{d} }&Vowel /a/   & Clinical setting\\

 MDVR-KCL \citep{jaeger2019mobile} & English & 16 & 21 &\ding{51}\textsuperscript{\href{https://data.niaid.nih.gov/resources?id=zenodo_2867215}{e}}& Sentences  & Real-world condition\\

Little \citep{parkinsons_174} & English & 23 & 8 &\ding{51}\textsuperscript{\href{https://archive.ics.uci.edu/dataset/174/parkinsons}{f} }& Vowel /a/
& Sound-treated booth\\

Oxford \citep{parkinsons_telemonitoring_189} & English & 42 & ---&\ding{51}\textsuperscript{\href{https://archive.ics.uci.edu/dataset/189/parkinsons+telemonitoring}{g} } & Vowel /a/ and sentences  &  Real-world condition\\

 \citet{iyer2023machine} & English & 40 & 41 &\ding{51}\textsuperscript{\href{https://figshare.com/articles/dataset/Voice_Samples_for_Patients_with_Parkinson_s_Disease_and_Healthy_Controls/23849127}{h} }& Vowel /a/ & Real-world condition\\

LSVT \citep{tsanas2012accurate} & English & 14 & --- &\ding{55} & Vowel /a/    & Sound attenuated room\\

m-Power \citep{bot2016mpower} & English & 1060 &5357 &\ding{55}& Vowel /a/   & Clinical setting\\

PC GITA~\citep{orozco2014new} & Spanish & 50 & 50 &\ding{55} & Vowels, speech, dialogues and sentences & Noise controlled \\

Naranjo \citep{parkinson_dataset_with_replicated_acoustic_features__489} & Spanish & 40 & 40 &\ding{51}\textsuperscript{\href{https://archive.ics.uci.edu/dataset/489/parkinson+dataset+with+replicated+acoustic+features}{i} }& Vowel /a/ & Clinical setting \\

\hline

BenSParX & Bengali & 60 & 60 &\ding{51}& Conversational speech in Bengali & Real-world condition \\
\bottomrule
\multicolumn{7}{l}{\textsuperscript{a} \url{https://archive.ics.uci.edu/dataset/392/early+biomarkers+of+parkinson+s+disease+based+on+natural+connected+speech}} \\
\multicolumn{7}{l}{\textsuperscript{b} {Sustained vowels (“aaa... ” and “eee...” in Chinese Pinyin)}}\\

\multicolumn{7}{l}{\textsuperscript{c} \url{https://archive.ics.uci.edu/dataset/470/parkinson+s+disease+classification}} \\

\multicolumn{7}{l}{\textsuperscript{d} \url{https://archive.ics.uci.edu/dataset/301/parkinson+speech+dataset+with+multiple+types+of+sound+recordings}} \\

\multicolumn{7}{l}{\textsuperscript{e} \url{https://data.niaid.nih.gov/resources?id=zenodo_2867215}}\\

\multicolumn{7}{l}{\textsuperscript{f} \url{https://archive.ics.uci.edu/dataset/174/parkinsons}} \\
\multicolumn{7}{l}{\textsuperscript{g} \url{https://archive.ics.uci.edu/dataset/189/parkinsons+telemonitoring}} \\

\multicolumn{7}{l}{\textsuperscript{h} \url{https://figshare.com/articles/dataset/Voice_Samples_for_Patients_with_Parkinson_s_Disease_and_Healthy_Controls/23849127}} \\

\multicolumn{7}{l}{\textsuperscript{i} \url{https://archive.ics.uci.edu/dataset/489/parkinson+dataset+with+replicated+acoustic+features}} \\

\end{tabular}%
}
\end{table}

Notable examples include datasets in Australian \citep{9887934,viswanathan2019complexity}, Italian \citep{dimauro2017assessment,scimeca2023robust,9650872}, Spanish \citep{parkinson_dataset_with_replicated_acoustic_features__489,orozco2014new}, Czech \citep{early_biomarkers_of_parkinsons_disease_based_on_natural_connected_speech_392,scimeca2023robust}, Korean \citep{jeong2024machine}, Mandarin {
\citep{wang2022early,ZhangTao59}}
, Portuguese \citep{proencca2014characterizing}, Turkish \citep{parkinsons_disease_classification_470,parkinsons_speech_with_multiple_types_of_sound_recordings_301}, and English \citep{jaeger2019mobile,parkinsons_174,parkinsons_telemonitoring_189,iyer2023machine,tsanas2012accurate}. These multilingual corpora have significantly advanced PD diagnosis by capturing language-specific acoustic characteristics. However, recent reviews {
\citep{xavier2025voice,NGO2022107133,zhao2025artificial,rabie2025review} }
highlight the absence of publicly available Bengali PD voice datasets, despite Bengali being the seventh most spoken language globally with over 230 million speakers. This gap limits inclusivity, restricts the study of language-specific markers, and hinders the development of generalizable diagnostic models.

Existing studies employ diverse speech modalities to capture different aspects of motor speech impairment. Many datasets {
\citep{orozco2014new,parkinson_dataset_with_replicated_acoustic_features__489,parkinsons_174,parkinsons_telemonitoring_189,iyer2023machine,tsanas2012accurate,parkinsons_disease_classification_470,parkinsons_speech_with_multiple_types_of_sound_recordings_301} }
rely on sustained vowels, which effectively capture phonatory irregularities such as tremor and dysphonia. Phoneme-level datasets in Australian English {
\citep{9887934,viswanathan2019complexity} }
enable fine-grained analysis of articulatory impairments. Other datasets {
\citep{dimauro2017assessment,scimeca2023robust,wang2022early,proencca2014characterizing,orozco2014new,parkinsons_telemonitoring_189} }
incorporate words and sentences to capture prosodic and temporal features. Notably, datasets such as PC-GITA {
\citep{orozco2014new} }
include spontaneous speech, which provides richer acoustic and prosodic information. Despite this diversity, only a limited number of datasets include natural conversational speech, which is crucial for ecologically valid PD detection. Conversational speech reflects real-world communication and introduces greater cognitive and motor demands, making subtle impairments more detectable. Studies such as {
\citep{vasquez2019convolutional} }
show that spontaneous dialogue improves discrimination between PD and healthy controls.%
The recording environment also affects dataset applicability. Many datasets {
\citep{9887934,viswanathan2019complexity,scimeca2023robust,jeong2024machine,proencca2014characterizing,parkinsons_disease_classification_470,parkinsons_speech_with_multiple_types_of_sound_recordings_301} }
are collected in controlled clinical settings, ensuring consistency but limiting real-world generalization. Models trained on such data often struggle in practical scenarios with noise and variability.

To address these limitations, we introduce BenSParX, the first publicly available Bengali conversational speech dataset for PD detection. Unlike prior datasets based on scripted speech in controlled environments, BenSParX captures spontaneous conversations recorded in real-world conditions, enhancing ecological validity. By incorporating Bengali, it addresses a major linguistic gap and supports the development of robust and generalizable AI-based diagnostic systems.

\subsection{Machine learning-based approaches}
Machine learning (ML) approaches for PD detection using handcrafted voice features exhibit substantial variation in methodological rigor -- particularly in terms of feature diversity, the application of feature selection techniques, hyperparameter tuning, and the integration of model explainability. A summary of representative studies is provided in Table~\ref{summary_of_related_works}.

\begin{table}[!ht]
\centering
\caption{Summary of machine learning-based PD detection approaches. \label{summary_of_related_works}}
 \resizebox{1\textwidth}{!}{%
\begin{tabular}{lllccccccccccl}
 \toprule
 \multirow{2}{*}{Study} & \multicolumn{2}{c}{Dataset} & \multicolumn{8}{c}{Feature} &  \multirow{2}{*}{\makecell[c]{Feature\\[-4pt]selection}}&\multirow{2}{*}{\makecell[c]{Explaina\\[-4pt]bility}}&   \multirow{2}{*}{\makecell[t l]{Best classifier}} \\
 \cmidrule(r){2-3} \cmidrule(r){4-11}
                & Name& Participant language   & MFCC & Jitter& Shimmer &  Harmonicity  &  Pulse & Formants &Pitch & Intensity  \\ \hline
 \midrule
\citep{9887934} &\citet{9887934}  & English &\ding{51}  & \ding{51} & \ding{51} & \ding{51} &\ding{55} &\ding{55} & \ding{55} &\ding{55}&\ding{51}&\ding{55}  &SVM\\

 \citep{swain2024towards}  &Little \citep{parkinsons_174}  &English  & \ding{55} & \ding{51} & \ding{51} & \ding{51}& \ding{55} &\ding{55} & \ding{55} & \ding{55} &\ding{55}&\ding{55} &KNN  \\

 \citep{9650872}  &\citet{9650872} & Italian&  \ding{51} & \ding{55} & \ding{55} & \ding{55}& \ding{55} &\ding{55} & \ding{55} & \ding{51} & \ding{51}& \ding{55} & SVM \\

 \citep{tsanas2012accurate}  &LSVT \citep{tsanas2012accurate} & English&\ding{51} & \ding{51} & \ding{51} & \ding{51}& \ding{55} &\ding{55} & \ding{51} & \ding{55} & \ding{51} &\ding{55} & SVM \\

  \citep{wang2022early} & \citet{wang2022early} & Mandarin& \ding{51} & \ding{51} & \ding{51} & \ding{55}& \ding{55} &\ding{55} & \ding{51} & \ding{51} &\ding{51}& \ding{55}  & SVM\\

 \citep{di2024machine}& MDVR-KCL \citep{jaeger2019mobile}   & English& \ding{51} & \ding{55} & \ding{55} & \ding{55}& \ding{55} &\ding{55} & \ding{55} & \ding{55} &\ding{55}&\ding{55} & KNN  \\

\citep{balaha2025comprehensive}  &MDVR-KCL \citep{jaeger2019mobile} &English& \ding{51} & \ding{51} & \ding{51} & \ding{55}& \ding{55} &\ding{55} & \ding{55} & \ding{55} &\ding{55} &\ding{55} &XGB \\

 \citep{toye2021comparative}  &MDVR-KCL \citep{jaeger2019mobile}  &English & \ding{51} & \ding{51} & \ding{51} & \ding{55}& \ding{55} &\ding{55} & \ding{55} & \ding{55} &\ding{51}& \ding{55} &SVM  \\
 \citep{braga2019automatic}&\citet{proencca2014characterizing}  & Portuguese& \ding{55} & \ding{51} & \ding{51} & \ding{51}& \ding{55} &\ding{55} & \ding{51} & \ding{55} &\ding{51}&\ding{55}    &RF  \\

 \citep{scimeca2023robust}& \citet{scimeca2023robust}  & Spanish, Italian, Czech  & \ding{51} & \ding{51} & \ding{51} & \ding{55}& \ding{55} &\ding{51} & \ding{55} & \ding{51} &\ding{51}& \ding{55}   & Ensemble  \\

 \citep{9426437}& PC GITA \citep{orozco2014new}  & Spanish & \ding{51} & \ding{51} & \ding{51} & \ding{55}& \ding{55} &\ding{55} & \ding{51} & \ding{51} &\ding{55}& \ding{55}  & SVM  \\

 \citep{mancini2024investigating}& PC GITA \citep{orozco2014new}  & Spanish & \ding{51} & \ding{51} & \ding{51} & \ding{55}& \ding{55} &\ding{55} & \ding{51} & \ding{51} &\ding{55}& \ding{51}  & CNN  \\

\citep{sakar2013collection}& Sakar13 \citep{parkinsons_speech_with_multiple_types_of_sound_recordings_301} & Turkish & \ding{55} & \ding{51} & \ding{51} & \ding{51}& \ding{51} &\ding{55} & \ding{51} & \ding{55} &\ding{55} & \ding{55}  & SVM \\

 \citep{10.1145/3441417.3441425} &Sakar18 \citep{parkinsons_disease_classification_470} & Turkish & \ding{51} & \ding{55} & \ding{55} & \ding{55}& \ding{55} &\ding{55} & \ding{55} & \ding{55} & \ding{55}&  \ding{55} &Stacking \\

 \citep{parisi2021m}  &Sakar18 \citep{parkinsons_disease_classification_470}, Little \citep{parkinsons_174}&English, Turkish & \ding{51} & \ding{51} & \ding{51} & \ding{51}& \ding{55} &\ding{51} & \ding{55} & \ding{51} &\ding{55} & \ding{55}&m-ark-SVM \\

\citep{singh2020robust}  &m-Power \citep{bot2016mpower}  &English & \ding{51} & \ding{55} & \ding{55} & \ding{55}& \ding{55} &\ding{55} & \ding{55} & \ding{55} &\ding{51}&  \ding{55} &SVM\\

\citep{shen2025explainable}  & \citet{iyer2023machine}  &English & \ding{51} & \ding{51} & \ding{51} & \ding{51}& \ding{55} &\ding{55} & \ding{51} & \ding{55} &\ding{55}&  \ding{51} &MLP+CNN+RNN+MKL\\

 \citep{NARANJO2017147}  & Naranjo \citep{parkinson_dataset_with_replicated_acoustic_features__489} &Spanish & \ding{51} & \ding{51} & \ding{51} & \ding{51}& \ding{55} &\ding{55} & \ding{51} & \ding{55} &\ding{51}& \ding{55} &Bayesian Binary Regression \\
 \citep{polat2020parkinson}  & Naranjo \citep{parkinson_dataset_with_replicated_acoustic_features__489} &Spanish & \ding{51} & \ding{51} & \ding{51} & \ding{51}& \ding{55} &\ding{55} & \ding{51} & \ding{55} &\ding{55}& \ding{55} &SVM\\

 \citep{jeong2024machine}  & \citet{jeong2024machine} &Korean  & \ding{51} & \ding{51} & \ding{51} & \ding{51}& \ding{55} &\ding{51} & \ding{51} & \ding{55} &\ding{55} & \ding{55}  &Soft voting \\

 \citep{ZHANG2021127}  & CPPDD \citep{ZhangTao59} &Mandarin  & \ding{55} & \ding{55} & \ding{55} & \ding{55}& \ding{55} &\ding{55} & \ding{55} & \ding{51} &\ding{55} & \ding{55}  &SVM \\

  \citep{lamba2023speech}  & ItalianPVS \citep{dimauro2017assessment} &Italian  & \ding{55} & \ding{51} & \ding{51} & \ding{55}& \ding{55} &\ding{55} & \ding{55} & \ding{55} &\ding{51} & \ding{55}  &LR \\

 \hline

 This study  &BenSParX  &  Bengali & \ding{51} & \ding{51} & \ding{51} & \ding{51}& \ding{51} &\ding{51} & \ding{51} & \ding{51} &\ding{51} & \ding{51} &RF \\
 \bottomrule

 \end{tabular}
 }
\end{table}

Among various voice features, Mel-Frequency Cepstral Coefficients (MFCCs) are the most widely used in PD voice classification {
\citep{9887934,tsanas2012accurate,wang2022early,di2024machine,balaha2025comprehensive,toye2021comparative,scimeca2023robust,9426437,mancini2024investigating,10.1145/3441417.3441425,nissar2019voice,yang2025optimizing,el2022enhancing,sakar2019comparative,8741725,parisi2021m,gunduz2019deep,gunduz2021efficient,singh2020robust,NARANJO2017147,polat2020parkinson,jeong2024machine}}
. Their effectiveness stems from their ability to capture perceptually relevant spectral characteristics and represent articulatory and phonatory deficits in PD. MFCCs are also robust to noise and effective in modeling short-term spectral variations critical for detecting dysarthric symptoms. Perturbation-based features such as jitter and shimmer are also widely used {
\citep{9887934,swain2024towards,balaha2025comprehensive,Hadjaidji_2025,wang2022early,toye2021comparative,scimeca2023robust,9426437,mancini2024investigating,10.1145/3441417.3441425,nissar2019voice,yang2025optimizing,sakar2013collection,NARANJO2017147,jeong2024machine,iyer2023machine,tsanas2012accurate,7042180}}
, as they quantify cycle-to-cycle variations in frequency and amplitude, reflecting irregular vocal fold vibrations commonly observed in PD. Harmonicity-related features, particularly Harmonics-to-Noise Ratio (HNR), are frequently employed \citep{9887934,swain2024towards,balaha2025comprehensive,Hadjaidji_2025,ecsa-11-20481,7042180,tsanas2012accurate,braga2019automatic,sakar2013collection,gunduz2019deep,gunduz2021efficient,NARANJO2017147,polat2020parkinson,jeong2024machine}, as they capture breathiness and hypophonia—key symptoms of PD. In contrast, pulse-related features remain underexplored due to the complexity of extraction, with limited studies \citep{sakar2013collection} utilizing them.

Prosodic features such as pitch and intensity are also incorporated in several studies {
\citep{wang2022early,ZHANG2021127,scimeca2023robust,9426437,mancini2024investigating,sakar2013collection,shen2025explainable,NARANJO2017147,polat2020parkinson,jeong2024machine,parisi2021m,gunduz2021efficient,gunduz2019deep}}
, as they help identify monotonic speech patterns such as monopitch and monoloudness. A smaller subset of works {
\citep{scimeca2023robust,sakar2019comparative,parisi2021m,gunduz2019deep,gunduz2021efficient} }
incorporates formants to capture articulatory imprecision. Despite this diversity, few studies perform a comprehensive analysis covering all major acoustic feature categories. Feature selection plays a crucial role in identifying the most discriminative and non-redundant features from high-dimensional data. However, only a limited number of studies {
\citep{9887934,ecsa-11-20481,9650872,tsanas2012accurate,wang2022early,scimeca2023robust,nissar2019voice,yang2025optimizing,el2022enhancing,sakar2019comparative,8741725} }
explicitly apply feature selection techniques. Most rely on a single predefined method, limiting robustness across datasets. As shown in Table~\ref{summary_of_related_works}, many studies use the full feature set without evaluating feature importance, which can introduce redundancy and reduce model generalizability.

Explainability is essential for clinical deployment, as it enhances transparency and trust by linking model predictions to meaningful acoustic features. However, only a few studies {
\citep{mancini2024investigating,yang2025optimizing,shen2025explainable} }
explicitly address this aspect by relating features such as jitter, shimmer, formants, and harmonicity to underlying physiological impairments. Most approaches focus primarily on performance, leaving explainability underexplored.

To address these limitations, the proposed framework integrates a comprehensive set of acoustic features, including underutilized formant and pulse-based measures. It employs a multi-stage feature selection pipeline, extensive hyperparameter tuning, and SHAP-based explainability to improve both performance and interpretability. This design supports accurate and transparent PD detection, particularly in low-resource and linguistically underrepresented settings.

\section{Methodology}
\begin{figure}[!ht]
\centering
\includegraphics[width=1\textwidth]{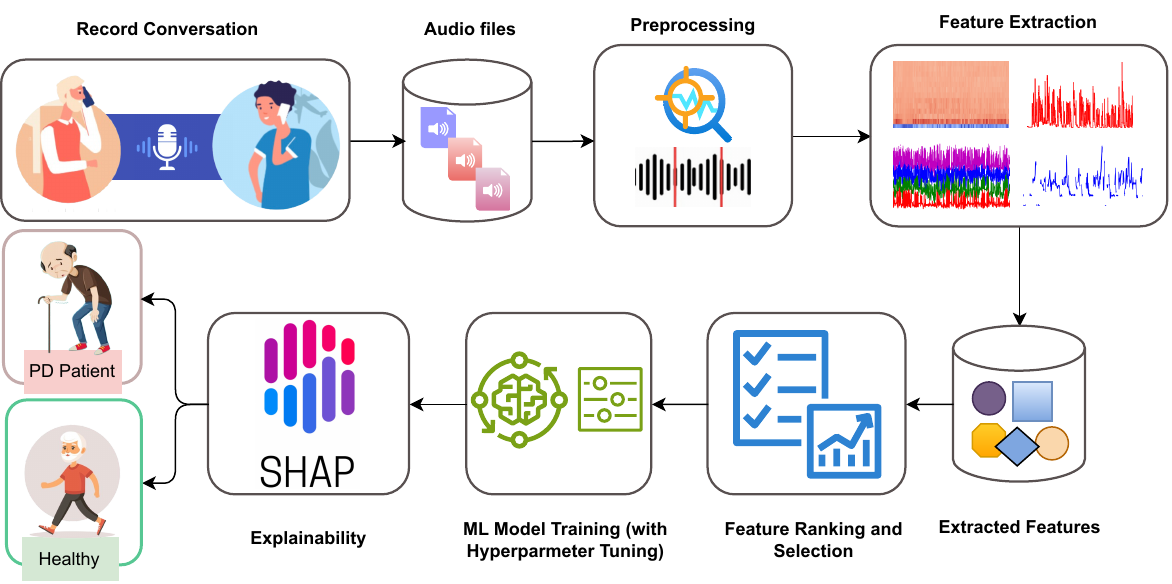}
\caption{A schematic overview of the framework for Parkinson's disease detection. \label{fig:methodology}}
\end{figure}

Figure~\ref{fig:methodology} illustrates a schematic diagram of our proposed framework, which is designed for robust and explainable PD detection from Bengali conversational speech. The methodology comprises several key stages: dataset collection and preprocessing, comprehensive acoustic feature extraction, systematic feature ranking and selection, extensive hyperparameter tuning, model training, and performance evaluation.

\subsection{Dataset collection}

We collaborated with the Movement Disorder Society of Bangladesh and Bangabandhu Sheikh Mujib Medical University (BSMMU) to develop a clinically validated dataset for PD detection. Patients diagnosed with PD within the past 1.5 years were recruited, ensuring confirmed medical diagnoses. Healthy control participants were recruited through public advertisements, with the inclusion criterion that they had no prior diagnosis of any neurodegenerative disorder, thereby establishing a reliable baseline for comparison. The study was conducted in accordance with the ethical principles outlined in the Declaration of Helsinki and received approval from the institutional review board (IRB) (approval number CSE201871). All participants provided informed consent prior to participation.

Data collection was conducted via phone call recordings using an Oppo F17 smartphone. Conversations followed a semi-structured interview format, where each participant, PD or healthy control (HC), was asked the same set of health-related questions: (i) Can you describe your current health condition? (ii) What medications are you currently taking? (iii) How often do you visit your doctor? and (iv) How has your condition affected your daily activities? These open-ended questions were designed to elicit natural, spontaneous speech and to ensure sufficient duration and variability for acoustic analysis.

Recordings were saved in uncompressed mono-channel WAV format with a sampling frequency of 44 kHz, ensuring high-fidelity audio capture suitable for detailed acoustic analysis. This setup preserves natural voice characteristics and minimizes processing artifacts. No audio compression or enhancement was applied.

The dataset comprises 120 recordings from 120 participants: 60 PD patients (male 40, female 20) and 60 healthy controls (male 35, female 25). The balanced sample composition supports robust comparative analysis between PD and control groups. Participants' demographic details are summarized in Table~\ref{tab:participant-demography}. The PD patients ranged in age from 43 to 65 years, and the healthy controls ranged from 41 to 57 years. Each participant contributed recordings of speech between 1 and 2 minutes. This length allows for intra-speaker variability, especially important in PD cases where vocal symptoms can fluctuate due to medication effects, fatigue, or disease progression.
\begin{table}[!ht]
\centering
\caption{Participant demographies. \label{tab:participant-demography}}
\begin{tabular}{lcc cccc c}
\toprule
\multirow{2}{*}{Participant} & \multicolumn{4}{c}{Age} & \multicolumn{2}{c}{Gender} & \multirow{2}{*}{Total}\\
 \cmidrule(r){2-5} \cmidrule(r){6-7}
            & Min & Max & Avg & SD & Male & Female & \\
\hline

\midrule
Parkinson's disease (PD) &43 & 65&59.4 & 10.38 & 40 & 20& 60\\
Healthy control (HC) &41 & 57& 50.5 & 6.15& 35 & 25 & 60\\
\bottomrule
\end{tabular}
\end{table}

\subsection{Data preprocessing}

\begin{figure}[!ht]
\centering
\includegraphics[width=1\textwidth]{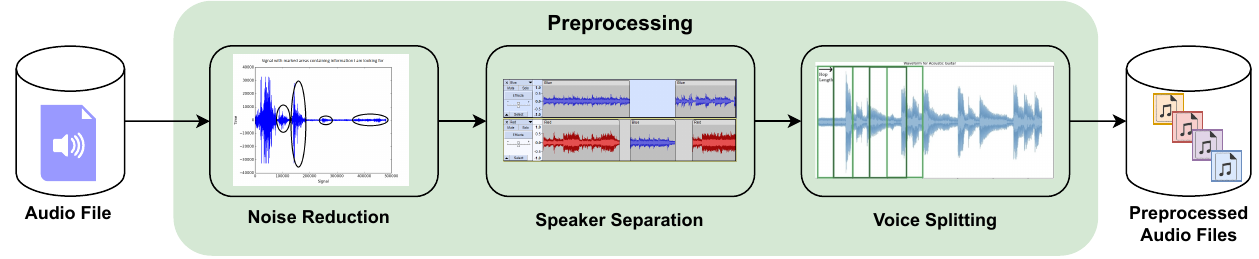}
\caption{Preprocessing steps for raw audio recordings. \label{fig:preprocessing}}
\end{figure}

Figure~\ref{fig:preprocessing} illustrates the audio preprocessing pipeline applied to the raw recordings. Given the real-world recording conditions, the audio data often included background noise such as ambient sounds, overlapping speech, and phone line interference. To ensure clean, speaker-specific input for feature extraction and model training, the preprocessing workflow consisted of three key stages: noise reduction, voice separation, and voice segmentation.

To mitigate background interference, we employed the noise reduction algorithm in Audacity~\citep{Audacity2021}. This method analyzes a user-specified segment containing only background noise to generate a noise profile, which is then used to attenuate noise throughout the full recording. This step preserves essential voice characteristics while reducing distortion from environmental factors.

Following denoising, voice separation was performed to isolate the participant's speech from that of the interviewer. Given the conversational nature of the recordings, this step was crucial to prevent contamination of feature analysis with non-target speech. Finally, each cleaned and isolated recording was segmented into 10-second intervals to facilitate efficient feature extraction. Depending on the duration of the original recording, this process yielded between five and eight segments per participant. In total, the segmentation step produced 900 high-quality audio samples from the 120 original recordings, as summarized in Table~\ref{tab:recordings-distribution}.

\begin{table}[!ht]
\centering
\caption{Distribution of the number of recordings per participant along with total sample counts for PD and HC groups.
\label{tab:recordings-distribution}}
\begin{tabular}{ccccc}
\hline
\multirow{2}{*}{Type} & \multicolumn{3}{c}{Voice samples per participant} & \multirow{2}{*}{Total voice samples} \\
 \cmidrule(r){2-4}    & Min & Max & Avg & \\ \hline \hline
PD   & 5       & 7       & 6       & 450 \\
HC   & 6       & 8       & 7       & 450 \\
\hline

\hline
\end{tabular}
\end{table}

\subsection{Feature extraction}
In our framework, we employed eight different categories of acoustic features as they are widely used in the literature for detecting PD (see Table~\ref{summary_of_related_works}): Spectrogram, i.e., mel-frequency cepstral coefficients (MFCCs), intensity, pitch, jitter, shimmer,  formants, pulse, and harmonicity.

MFCCs are widely used in speech processing due to their strong alignment with human auditory perception. They provide a compact representation of the spectral envelope, emphasizing perceptually important frequency components. In this study, we extracted 13 base MFCCs (coefficients 0--12), along with their first and second-order derivatives ($\Delta$ and $\Delta^2$), resulting in a total of 39 MFCC-based features per segment. These features capture both spectral characteristics and temporal dynamics, which are essential for identifying instability in Parkinsonian speech.

Intensity reflects the loudness or energy of a speech signal and is important for capturing prosodic variation and vocal effort, both of which are often reduced in PD. We extracted seven intensity-related features: \texttt{minimum intensity}, \texttt{maximum intensity}, \texttt{mean intensity}, \texttt{mean log energy}, and the first-, second-, and third-order derivatives of log energy ($\Delta$, $\Delta^2$, $\Delta^3$), enabling characterization of both static energy levels and temporal variations.

Pitch (fundamental frequency, F0) represents vocal fold vibration rate and is crucial for analyzing intonation and prosodic control. Impairments in PD often lead to reduced pitch variation and instability. We extracted two pitch features: \texttt{minimum pitch} and \texttt{maximum pitch}, which together describe the vocal range and pitch modulation capability.

Jitter measures cycle-to-cycle variations in fundamental frequency and serves as an indicator of pitch instability caused by irregular vocal fold vibration. To comprehensively capture these variations, we extracted five jitter features: \texttt{Jitter (Local)}, \texttt{Jitter (Local Absolute)}, \texttt{RAP}, \texttt{PPQ5}, and \texttt{DDP}.

Shimmer quantifies cycle-to-cycle amplitude variations and reflects instability in vocal intensity, often perceived as hoarseness or breathiness in PD speech. We extracted six shimmer features: \texttt{Shimmer (Local)}, \texttt{Shimmer (Local dB)}, \texttt{APQ3}, \texttt{APQ5}, \texttt{APQ11}, and \texttt{DDA}, enabling multi-scale analysis of amplitude perturbations.

Formants represent the resonant frequencies of the vocal tract and are critical for characterizing articulatory behavior. Changes in formant patterns can indicate reduced articulatory precision in PD. We extracted the first four formants (\texttt{F1}--\texttt{F4}) to capture these articulatory characteristics.

Pulse-based features provide insight into the temporal regularity of vocal fold vibrations and are useful for assessing voice stability. We extracted four pulse-related features: \texttt{number of pulses}, \texttt{number of periods}, \texttt{mean period}, and \texttt{standard deviation of period durations}.

Harmonicity measures the balance between periodic (harmonic) and aperiodic (noise) components in speech, reflecting voice quality and phonatory stability. Lower harmonicity is commonly associated with PD. We extracted four harmonicity features: \texttt{AutoCorrHarmonicity}, \texttt{mean AutoCorrHarmonicity}, \texttt{HNR}, and \texttt{NHR}. A comprehensive description of the feature extraction pipeline and acoustic feature formulations is provided in the Supplementary Material.

Table~\ref{tab:categorized_pd_hc_features} reports the statistical differences between PD and HC groups across various acoustic feature categories with statistical significance. Significant differences were observed in pulse analysis and spectrogram features, i.e., MFCCs, with $p<0.001$, indicating strong group-level distinctions in speech rhythm and spectral energy patterns. Jitter, shimmer, intensity, and formants-related features also showed high discriminative power ($p<0.001$), reflecting instability in frequency, amplitude, and loudness—key markers of hypokinetic dysarthria. Harmonicity (HNR) and pitch features demonstrated moderate significance ($p<0.05$). These findings validate the relevance of a multi-feature approach in capturing the complex acoustic manifestations of PD.

{\renewcommand{\arraystretch}{.8}%
\begin{longtable}[c]{llccc}
    \caption{Extracted feature values for PD and HC groups} \label{tab:categorized_pd_hc_features} \\

    \toprule
    Category & Feature & PD (Mean $\pm$ Std) & HC (Mean $\pm$ Std) & \textit{p}-value\\ \hline\hline
    \endfirsthead

    \multicolumn{4}{l}{\textit{{\tablename\ \thetable{} -- continued from previous page}}}\\
    \toprule
   Category & Feature & PD (Mean $\pm$ Std) & HC (Mean $\pm$ Std) & \textit{p}-value\\ \hline
    \endhead

     \hline
    \endfoot

    \hline

    \hline
    \endlastfoot

    Pulse
    & Number of Pulses & 325.8530 $\pm$ 92.3126 & 398.5797 $\pm$ 118.2790 & \multirow{4}{*}{$<0.001$}\\
    & Number of Period Pulses & 324.4074 $\pm$ 92.7374 & 397.2649 $\pm$ 117.9763 &\\
    & Mean Period Pulses & 0.0071 $\pm$ 0.0018 & 0.0060 $\pm$ 0.0021 &\\
    & Standard Deviation Period Pulses & 0.0005 $\pm$ 0.0008 & 0.0003 $\pm$ 0.0006 &\\

    \hline
    Spectrogram
    & MFCC0 & 1.8790 $\pm$ 0.2988 & 2.0082 $\pm$ 0.2409 & \multirow{39}{*}{$<0.001$}\\
    & MFCC1 & 0.8059 $\pm$ 0.2231 & 0.8265 $\pm$ 0.1748 \\
    & MFCC2 & 0.2148 $\pm$ 0.1853 & 0.0190 $\pm$ 0.2233 \\
    & MFCC3 & 0.0554 $\pm$ 0.2101 & -0.0580 $\pm$ 0.1786 \\
    & MFCC4 & -0.1673 $\pm$ 0.1633 & -0.2134 $\pm$ 0.1519 \\
    & MFCC5 & -0.0906 $\pm$ 0.1307 & -0.1380 $\pm$ 0.1472 \\
    & MFCC6 & -0.1567 $\pm$ 0.1142 & -0.1374 $\pm$ 0.0837 \\
    & MFCC7 & 0.0243 $\pm$ 0.1025 & -0.0156 $\pm$ 0.1081 \\
    & MFCC8 & 0.0554 $\pm$ 0.1096 & 0.0389 $\pm$ 0.0937 \\
    & MFCC9 & 0.0172 $\pm$ 0.1054 & -0.0107 $\pm$ 0.1024 \\
    & MFCC10 & -0.0893 $\pm$ 0.0928 & -0.0858 $\pm$ 0.0989 \\
    & MFCC11 & -0.0380 $\pm$ 0.0794 & -0.0481 $\pm$ 0.0656 \\
    & MFCC12 & -0.0137 $\pm$ 0.0808 & -0.0226 $\pm$ 0.0778 \\
    & $\Delta$MFCC0 & 15.6582 $\pm$ 2.4897 & 16.7351 $\pm$ 2.0078 \\
    & $\Delta$MFCC1 & 6.7155 $\pm$ 1.8591 & 6.8877 $\pm$ 1.4565 \\
    & $\Delta$MFCC2 & 1.7899 $\pm$ 1.5438 & 0.1582 $\pm$ 1.8607 \\
    & $\Delta$MFCC3 & 0.4616 $\pm$ 1.7504 & -0.4833 $\pm$ 1.4880 \\
    & $\Delta$MFCC4 & -1.3941 $\pm$ 1.3606 & -1.7783 $\pm$ 1.2658 \\
    & $\Delta$MFCC5 & -0.7549 $\pm$ 1.0892 & -1.1502 $\pm$ 1.2266 \\
    & $\Delta$MFCC6 & -1.3059 $\pm$ 0.9519 & -1.1452 $\pm$ 0.6974 \\
    & $\Delta$MFCC7 & 0.2021 $\pm$ 0.8540 & -0.1299 $\pm$ 0.9009 \\
    & $\Delta$MFCC8 & 0.4614 $\pm$ 0.9133 & 0.3239 $\pm$ 0.7807 \\
    & $\Delta$MFCC9 & 0.1434 $\pm$ 0.8782 & -0.0891 $\pm$ 0.8536 \\
    & $\Delta$MFCC10 & -0.7442 $\pm$ 0.7732 & -0.7150 $\pm$ 0.8238 \\
    &$\Delta$MFCC11 & -0.3166 $\pm$ 0.6613 & -0.4005 $\pm$ 0.5469 \\
    & $\Delta$MFCC12 & -0.1143 $\pm$ 0.6731 & -0.1882 $\pm$ 0.6483 \\
    & $\Delta^2$ MFCC0 & 1.0122 $\pm$ 0.6946 & 0.7779 $\pm$ 0.3315 \\
    &$\Delta^2$MFCC1 & 0.6321 $\pm$ 0.3563 & 0.5357 $\pm$ 0.1895 \\
    & $\Delta^2$MFCC2 & 0.5024 $\pm$ 0.1916 & 0.4798 $\pm$ 0.1680 \\
    & $\Delta^2$MFCC3 & 0.4575 $\pm$ 0.1815 & 0.4080 $\pm$ 0.1377 \\
    & $\Delta^2$MFCC4 & 0.4160 $\pm$ 0.1545 & 0.3549 $\pm$ 0.0895 \\
    &$\Delta^2$MFCC5 & 0.3854 $\pm$ 0.1382 & 0.3391 $\pm$ 0.1101 \\
    & $\Delta^2$MFCC6 & 0.3834 $\pm$ 0.1297 & 0.3238 $\pm$ 0.0900 \\
    & $\Delta^2$MFCC7 & 0.3434 $\pm$ 0.1032 & 0.2866 $\pm$ 0.0785 \\
    & $\Delta^2$MFCC8 & 0.3200 $\pm$ 0.0908 & 0.2674 $\pm$ 0.0750 \\
    & $\Delta^2$MFCC9 & 0.3035 $\pm$ 0.0824 & 0.2673 $\pm$ 0.0757 \\
    & $\Delta^2$MFCC10 & 0.3002 $\pm$ 0.0811 & 0.2561 $\pm$ 0.0640 \\
    & $\Delta^2$MFCC11 & 0.2847 $\pm$ 0.0781 & 0.2474 $\pm$ 0.0632 \\
    &$\Delta^2$MFCC12 & 0.2698 $\pm$ 0.0704 & 0.2413 $\pm$ 0.0691 \\

    \hline%
    Jitter
    & Local & 0.0029 $\pm$ 0.0033 & 0.0014 $\pm$ 0.0012 & \multirow{5}{*}{$<0.001$}\\
    & Local Absolute & 0.000022 $\pm$ 0.000029 & 0.000009 $\pm$ 0.00001 \\
    & RAP & 0.0008 $\pm$ 0.0013 & 0.0003 $\pm$ 0.0004 \\
    & PPQ5 & 0.0015 $\pm$ 0.0022 & 0.0006 $\pm$ 0.0007 \\
    & DDP & 0.0024 $\pm$ 0.0038 & 0.0009 $\pm$ 0.0012 \\

    \hline
    Shimmer
    & Local & 0.0775 $\pm$ 0.0465 & 0.0582 $\pm$ 0.0426 & \multirow{6}{*}{$<0.001$} \\
    & Local dB & 0.6987 $\pm$ 0.4114 & 0.5148 $\pm$ 0.3764 \\
    & APQ3 & 0.0390 $\pm$ 0.0243 & 0.0309 $\pm$ 0.0228 \\
    & APQ5 & 0.0472 $\pm$ 0.0299 & 0.0359 $\pm$ 0.0263 \\
    & APQ11 & 0.0649 $\pm$ 0.0383 & 0.0443 $\pm$ 0.0292 \\
    &DDA    & 0.1171 $\pm$ 0.0729 &     0.0927 $\pm$ 0.0684\\

    \hline
    Harmonicity
    & Auto CorrHarmonicity & 1.0218 $\pm$ 0.0774 & 1.0390 $\pm$ 0.0564 & \multirow{4}{*}{$<0.05$}\\
    & Mean Auto CorrHarmonicity & 0.0655 $\pm$ 0.1294 & 0.0347 $\pm$ 0.0796 \\
    & NHR & 19.3323 $\pm$ 6.0655 & 22.4007 $\pm$ 5.4898 \\
    & HNR & 3.2300 $\pm$ 0.0469 & 3.2313 $\pm$ 0.0893 \\

    \hline
    Intensity
    & Minimum Intensity & 74.2125 $\pm$ 9.8722 & 80.0942 $\pm$ 5.4893 & \multirow{7}{*}{$<0.001$}\\
    & Maximum Intensity & 80.4993 $\pm$ 7.4376 & 84.0747 $\pm$ 3.2203 \\
    & Mean Intensity & 78.0001 $\pm$ 8.0731 & 82.3882 $\pm$ 3.4014 \\
     & Mean Log Energy & 10.4472 $\pm$ 0.6014 & 10.7171 $\pm$ 0.4832 \\
   & $\Delta$ Log Energy      & 0.3366 $\pm$ 0.3149 &     0.2075 $\pm$ 0.1592\\
      &$\Delta^2$ Log Energy    &  0.0446 $\pm$ 0.0310      &0.0258 $\pm$  0.0209\\
& $\Delta^3$ Log Energy  &   0.0167 $\pm$ 0.0101 &     0.0092 $\pm$ 0.0060\\

    \hline
    Formants
    & f1 & 664.0428 $\pm$ 143.2074 & 758.2389 $\pm$ 136.4111 &\multirow{4}{*}{$<0.001$}\\
    & f2 & 1297.5350 $\pm$ 163.6211 & 1380.9154 $\pm$ 205.5758 \\
    & f3 & 2757.5436 $\pm$ 423.9698 & 2681.5449 $\pm$ 428.5507 \\
    & f4 & 3809.6557 $\pm$ 378.0670 & 3772.0340 $\pm$ 399.9208 \\

    \hline
   Pitch
    & Maximum Pitch & 37.6959 $\pm$ 34.1547 & 24.1527 $\pm$ 28.9559 & \multirow{2}{*}{$<0.05$}\\
    & Minimum Pitch & 1.3406 $\pm$ 0.0889 & 1.3813 $\pm$ 0.0979 \\
\end{longtable}
}%

\begin{figure}[!ht]
    \centering
    \begin{subfigure}[b]{0.32\linewidth}
        \centering
        \includegraphics[width=\textwidth]{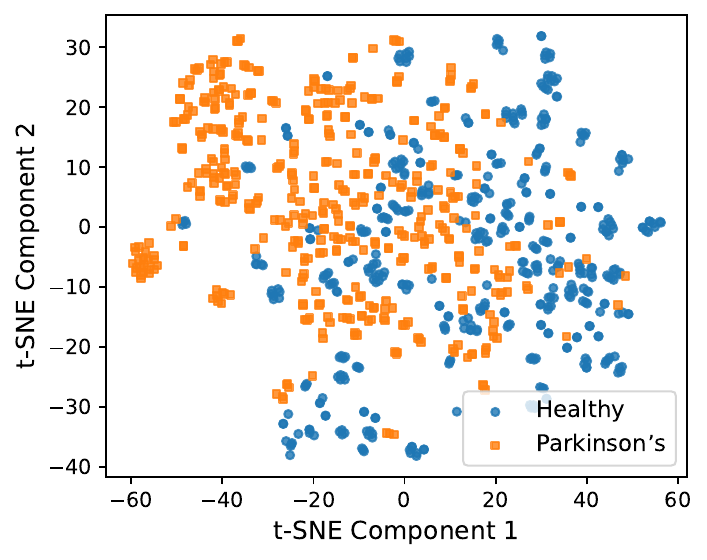}
        \caption{2D t-SNE scatter plot.}
        \label{fig:tsne_scatter}
    \end{subfigure}
    \hfill
    \begin{subfigure}[b]{0.32\linewidth}
        \centering
        \includegraphics[width=\textwidth]{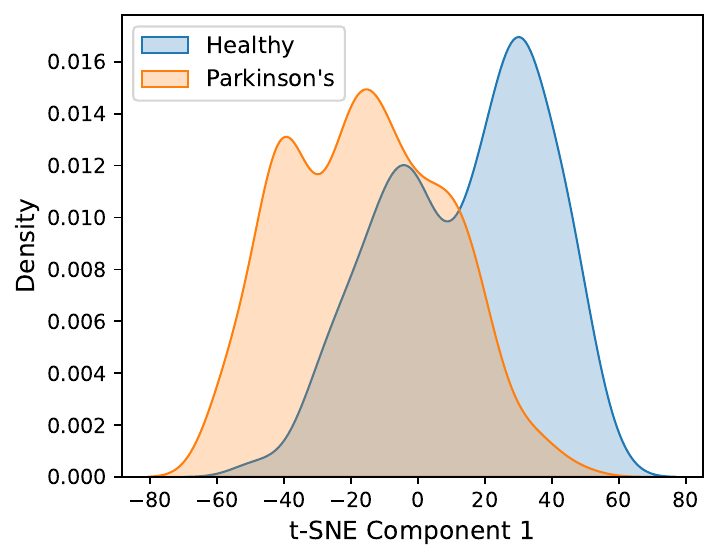}
        \caption{KDE plot for the first t-SNE component.}
        \label{fig:tsne1_dist}
    \end{subfigure}
    \hfill
    \begin{subfigure}[b]{0.32\linewidth}
        \centering
        \includegraphics[width=\textwidth]{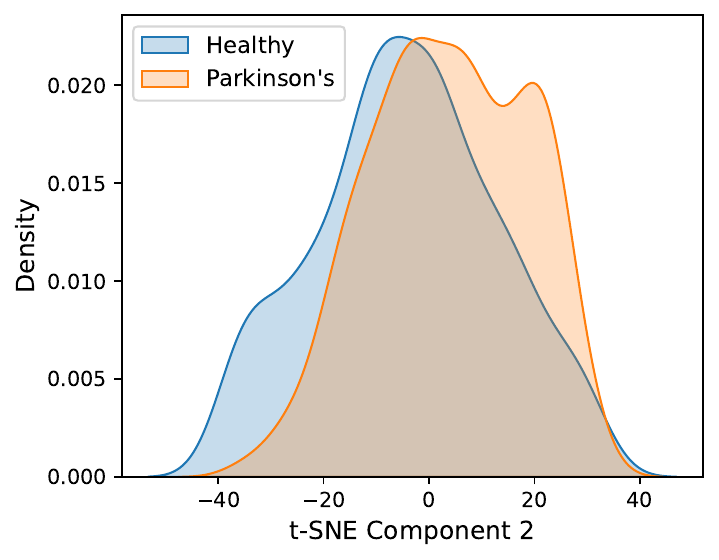}
        \caption{KDE plot for the second t-SNE component.}
        \label{fig:tsne2_dist}
    \end{subfigure}

    \caption{Visualization of \textit{t}-SNE results showing the distribution of Healthy and Parkinson's samples after dimensionality reduction: (a) 2D scatter plot of \textit{t}-SNE components, (b) KDE plot along the first \textit{t}-SNE component, and (c) KDE plot along the second \textit{t}-SNE component. These visualizations highlight the separability between the two classes, particularly along the first \textit{t}-SNE component.}
    \label{fig:tsne}
\end{figure}

Figure \ref{fig:tsne} illustrates the low-dimensional representation of the extracted features obtained through \textit{t}-distributed Stochastic Neighbor Embedding (\textit{t}-SNE). In Figure \ref{fig:tsne_scatter}, a two-dimensional scatter plot demonstrates that samples corresponding to PD and HC form distinguishable clusters, suggesting that the extracted features possess significant discriminative ability. Figure~\ref{fig:tsne1_dist} presents the kernel density estimation (KDE) plot along the first \textit{t}-SNE component, where a noticeable shift in the distributions between the two classes is evident, indicating that this component captures substantial inter-class variability. Figure~\ref{fig:tsne2_dist} shows the KDE plot along the second \textit{t}-SNE component, where moderate overlap is observed, yet a slight difference between the class distributions persists. Collectively, these visualizations confirm that the feature space effectively differentiates between PD and HC samples, with the first \textit{t}-SNE component contributing more prominently to the observed separability.

\subsection{Feature selection}

Given the high dimensionality and strong correlation of acoustic speech features, an effective feature selection method is essential for improving classification performance, reducing redundancy, and enhancing model interpretability in PD detection. In this study, we adopted four widely used and complementary feature selection methods: Recursive Feature Elimination with Cross-Validation (RFECV), Least Absolute Shrinkage and Selection Operator (LASSO), Relief-F combined with Sequential Forward Selection (SFS), and the Mann-Whitney U test.

These methods were adopted because they represent distinct selection philosophies widely validated in recent biomedical and speech processing literature: wrapper-based optimization (RFECV), embedded regularization (LASSO), filter-based ranking with greedy search (Relief-F+SFS), and non-parametric statistical filtering (Mann-Whitney U). The combination of these approaches ensures both statistical robustness and model-specific optimization, which has been shown to improve generalization in the PD speech analysis task.

Overall, integrating these heterogeneous feature selection methods enables a comprehensive evaluation of feature relevance from statistical, model-based, and optimization perspectives, which is critical for robust PD detection from speech signals. Further details on the feature selection pipeline are available in the Supplementary Material.

\subsection{Machine learning classifiers}

To build a robust PD detection system, the proposed framework integrates nine machine learning classifiers that have been consistently validated in prior speech-based PD studies. Instead of relying on a single best model, we intentionally combine diverse learning paradigms to capture both linear and nonlinear characteristics of acoustic features. The selection of each model is guided by empirical evidence from the literature and its complementary role within our framework.

\paragraph{Logistic regression (LR)}
LR has been widely adopted in PD speech classification due to its strong generalization and stability in high-dimensional feature spaces~\citep{swain2024towards,nissar2019voice,el2022enhancing,8741725,balaha2025comprehensive,ecsa-11-20481,toye2021comparative,sakar2019comparative}. Notably, \citet{8741725} reported top performance using LR, indicating that PD-related acoustic features often exhibit meaningful linear separability. Motivated by this, we include LR in our framework as a reliable baseline to validate whether the extracted features alone can discriminate PD without requiring complex nonlinear modeling.

\paragraph{Decision tree (DT)}
DT has been frequently used in PD detection due to its interpretability and ability to capture nonlinear relationships~{
\citep{balaha2025comprehensive, ecsa-11-20481, toye2021comparative, swain2024towards, scimeca2023robust, nissar2019voice, 10.1145/3441417.3441425}}
. Its transparent structure allows researchers to identify which acoustic features contribute most to classification decisions. In our study, DT is included not only for classification but also to provide interpretable insights into feature behavior, which is important for clinical relevance.

\paragraph{Random forest (RF)}
RF has consistently demonstrated strong performance in PD speech analysis by reducing overfitting and capturing complex feature interactions~{
\citep{swain2024towards,9650872,braga2019automatic,scimeca2023robust,nissar2019voice,el2022enhancing,8741725,balaha2025comprehensive,ecsa-11-20481,toye2021comparative,sakar2019comparative}}
. Studies such as \citet{el2022enhancing} and \citet{braga2019automatic} highlight its robustness across datasets. Based on this evidence, RF is used in our framework as both a high-performing classifier and a core component of the feature selection pipeline (RFECV), leveraging its stable feature importance estimation.

\paragraph{Support vector machine (SVM)}
SVM is one of the most extensively used classifiers in PD detection and often achieves superior accuracy due to its ability to handle high-dimensional and nonlinear data~{
\citep{9887934,9650872,9426437,sakar2013collection,nissar2019voice,toye2021comparative,parisi2021m,sakar2019comparative}}
. Prior studies consistently report strong sensitivity and robustness, particularly with kernel-based extensions. This motivates its inclusion in our framework as a benchmark model for capturing complex decision boundaries in acoustic space.

\paragraph{K-nearest neighbors (KNN)}
KNN has shown competitive performance in several PD detection studies, particularly in capturing subtle local variations in speech features~\citep{8741725,balaha2025comprehensive,Hadjaidji_2025,ecsa-11-20481,toye2021comparative,sakar2019comparative,swain2024towards,9650872,wang2022early,nissar2019voice,sakar2013collection}. For instance, {
\citet{swain2024towards} }
and {
\citet{wang2022early} }
demonstrated that KNN can outperform more complex models in certain scenarios. We include KNN in our framework to exploit these local feature relationships and to provide a complementary perspective to global learning models.

\paragraph{Multi-layer perceptron (MLP)}
MLP has been explored in PD detection for its ability to learn nonlinear mappings between acoustic features and disease states~\citep{ecsa-11-20481, sakar2019comparative, 9426437, nissar2019voice}. Notably, \citet{ecsa-11-20481} reported the best performance using MLP among several models. Based on this, MLP is incorporated in our framework to assess whether deeper nonlinear representations can further enhance classification beyond traditional ML approaches.

\paragraph{Gradient boosting (GB)}
GB has gained attention in PD classification for its strong predictive capability and ability to iteratively focus on hard-to-classify samples~{
\citep{balaha2025comprehensive,toye2021comparative,scimeca2023robust}}
. These properties make it particularly effective in handling subtle and noisy patterns in speech data. In our framework, GB is used as a high-accuracy model that complements RF and XGBoost in capturing complex feature interactions.

\paragraph{Extreme gradient boosting (XGBoost)}
XGBoost has been widely reported as one of the top-performing models for structured biomedical data, including PD detection~\citep{nissar2019voice, yang2025optimizing, balaha2025comprehensive}. Its regularization mechanism and efficient optimization make it robust against overfitting while maintaining high accuracy. In the proposed framework, XGBoost is used both as a classifier and as a feature ranking tool, leveraging its built-in importance scoring.

\paragraph{Adaptive boosting (AdaBoost)}
AdaBoost has been applied in PD detection to improve classification by focusing on misclassified samples~{
\citep{scimeca2023robust, 10.1145/3441417.3441425, balaha2025comprehensive}}
. Although not always the top performer, prior studies show that it enhances robustness, particularly in noisy and imbalanced datasets. In our framework, AdaBoost is included to increase model diversity and to evaluate performance stability across different learning strategies.

\subsection{Model training and evaluation}
\label{sec:explainability}

All models were trained and evaluated using a reproducible and leakage-free experimental pipeline. To ensure subject-level independence, we employed stratified group-wise cross-validation, where grouping was performed based on speaker ID. This guarantees that recordings from the same individual do not appear in both train and test sets, thereby ensuring that model evaluation remains unbiased and representative of unseen subjects.

A nested cross-validation framework was employed to obtain unbiased estimates of model performance while preventing information leakage. The outer loop consisted of 10-fold cross-validation for performance evaluation, while an inner 5-fold cross-validation was used for hyperparameter tuning and feature selection within the training data of each outer fold.
Specifically, for each outer fold, the data were partitioned into a training set (90\%) and an independent test set (10\%). Within the training set, hyperparameter optimization and feature selection were performed using 5-fold cross-validation. These procedures were implemented as part of a unified modeling pipeline to ensure that feature selection was conducted exclusively within the training data of each inner split. The same fold-wise models and held-out predictions from this nested cross-validation procedure were subsequently reused for the SHAP-based explainability analysis, ensuring that feature attributions, like performance metrics, are derived exclusively from data unseen during training.
Following inner cross-validation, the optimal hyperparameters and feature selection configuration were re-fitted on the entire outer training set. The resulting model was then evaluated on the held-out outer test fold, yielding out-of-fold predictions for each sample. This process was repeated across all 10 folds such that each observation received exactly one prediction from a model that was not trained on that observation.

Performance metrics were computed for each outer test fold and averaged to obtain overall estimates of generalization performance. To quantify uncertainty, 95\% confidence intervals were derived using stratified nonparametric bias-corrected and accelerated (BCa) bootstrap resampling with 10,000 repetitions applied to the pooled out-of-fold predictions. All procedures were conducted with a fixed random seed (42) to ensure reproducibility, and stratified sampling was used where applicable. All models and feature selection methods were evaluated under identical cross-validation splits.

\subsection{Explainability}
\label{sec:explan}
Explainability is a critical component in the development and deployment of ML models for clinical decision support. It enhances transparency, builds trust among healthcare professionals, and supports clinical validation by clarifying how input features influence model predictions. In the context of PD detection from voice recordings, explainability ensures that model predictions can be interpreted regarding known clinical symptoms and acoustic biomarkers, thereby enhancing their real-world utility.

Several interpretability techniques have been proposed to address this need, including SHAP (SHapley Additive exPlanations), LIME (Local Interpretable Model-agnostic Explanations), and permutation importance. These methods assign importance scores to input features, helping to elucidate how individual features contribute to a model's output. Among them, SHAP stands out due to its strong theoretical foundation in cooperative game theory, offering both global and local interpretability. This dual capability makes SHAP particularly appropriate for clinical settings where understanding both general model behavior and individual patient predictions is essential.

 Recent studies on PD detection from speech, such as \citet{mancini2024investigating}, \citet{yang2025optimizing}, and \citet{shen2025explainable}, have applied SHAP to reveal the contribution of specific acoustic features to classification outcomes. Building on this precedent, we integrate SHAP into the same nested cross-validation procedure described in Section \ref{sec:explainability}, rather than fitting a separate model on the full dataset. Specifically, within each of the 10 outer folds, Mann-Whitney U feature selection and best  machine learning model with hyperparameter tuning are performed using only that fold's training partition, and SHAP values are computed using a TreeExplainer built on the resulting fold-specific model, applied exclusively to that fold's held-out test partition. Because each sample is held out exactly once across the 10 outer folds, concatenating these values yields one out-of-fold SHAP value per feature per sample for the entire dataset, with no leakage between feature attribution and model training. This design ensures that reported feature importances, like the performance metrics in Table \ref{tab:performance-comparison}, reflect only held-out predictive behavior. By doing so, we aim to bridge the gap between predictive performance and clinical interpretability, advancing the development of transparent, explainable, and trustworthy AI-driven tools for early PD diagnosis.

\subsection{Evaluation metrics}
To evaluate the performance of the trained classifiers, we employed standard confusion matrix-based metrics, including accuracy, precision, recall, specificity, F1 score, and AUC-ROC (Area Under the Receiver Operating Characteristic Curve). Accuracy reflects the overall correctness of the model by quantifying the proportion of all correctly classified instances. Precision (positive predictive value) measures the proportion of true positives among all predicted positives, indicating the reliability of the model's positive predictions. Recall (or sensitivity) captures the model's ability to correctly identify all actual positive cases. F1 score combines precision and recall into a single metric, offering a balanced evaluation, especially in the presence of class imbalance. Specificity measures the proportion of actual negative cases correctly identified, offering a counterbalance to recall.

In addition to these metrics, we also report the AUC-ROC, which evaluates a model's ability to discriminate between classes across all threshold settings. AUC-ROC provides a threshold-independent measure of separability, where a value of 1.0 indicates perfect classification, and 0.5 represents random guessing. This metric is particularly relevant in medical diagnostics, where sensitivity and specificity must be jointly optimized.

Furthermore, to assess the quality of probabilistic predictions, we incorporate the Brier score, a strictly proper scoring rule that measures the mean squared difference between predicted probabilities and actual binary outcomes. Unlike accuracy or AUC-ROC, which primarily evaluate classification performance, the Brier loss captures both discrimination and calibration aspects of a model. Lower Brier loss indicates better-calibrated and more reliable probability estimates, which is especially important in clinical applications where decision-making often depends on predicted risk probabilities rather than hard class labels.

Each model was trained and validated using 10-fold cross-validation, and the resulting metrics were computed on the held-out test folds to ensure robust and generalizable performance estimates. The metrics are formally defined as follows:
\begin{equation}
\text{Accuracy} = \frac{TP + TN}{TP + TN + FP + FN}
\end{equation}
\begin{equation}
\text{Precision} = \frac{TP}{TP + FP}
\end{equation}
\begin{equation}
\text{Recall} = \frac{TP}{TP + FN}
\end{equation}
\begin{equation}
\text{F1 Score} = \frac{2 \times \text{Precision} \times \text{Recall}}{\text{Precision} + \text{Recall}}
\end{equation}
\begin{equation}
\text{Brier loss} = \frac{1}{N} \sum_{i=1}^{N} (p_i - y_i)^2
\end{equation}
Here, $TP$, $TN$, $FP$, and $FN$ represent the number of
true positives, false positives, true negatives, and false negatives, respectively. In the Brier score formulation, $p_i$ denotes the predicted probability of the positive class (i.e., PD) for instance $i$, $y_i \in {0,1}$ represents the true label, and $N$ is the total number of samples.

These metrics collectively offer a comprehensive assessment of classifier performance, ensuring balanced evaluation of both detection power and reliability, which is critical in the context of PD screening.

\section{Results}

\subsection{Best features}
Table~\ref{tab:featureselection} presents the distribution of selected features across different acoustic categories using various combinations of ranking techniques and feature selection methods within the proposed framework. The results are disaggregated across eight acoustic categories, \texttt{Pulse}, \texttt{MFCCs}, \texttt{Jitter}, \texttt{Shimmer}, \texttt{Harmonicity}, \texttt{Intensity}, \texttt{Formants}, and \texttt{Pitch}. This table reports the range (minimum--maximum) of the number of selected features across the 10 outer cross-validation folds, reflecting variability in feature selection across different training partitions.
\begin{table}[!ht]
\caption{Best feature selection using different feature selection methods.
\label{tab:featureselection}}
\centering
\resizebox{\textwidth}{!}{
\begin{tabular}{ll ccccccccc}

\hline
\multirow{2}{*}{\makecell{Feature\\importance}} &
\multirow{2}{*}{\makecell[t l]{Selection\\method}} &
\multicolumn{8}{c}{Feature category} &
\multirow{2}{*}{\makecell{Total\\(n=71)}} \\
\cline{3-10}
& & \makecell{Pulse\\(n=4)}& \makecell{MFCC\\(n=39)} & \makecell{Jitter\\(n=5)} & \makecell{Shimmer\\(n=6)} & \makecell{Harmonicity\\(n=4)} & \makecell{Intensity\\(n=7)} & \makecell{Formants\\(n=4)} & \makecell{Pitch\\(n=2)} & \\
\hline
\hline

\multirow{9}{*}{Relief-F}
 & SFS+AdaBoost     & 0--3 & 2--14 & 0--1 & 0--1 & 1--3  &1--4   & 0--2  & 0--2  & 15--24 \\
 & SFS+DT          & 0--3 & 5--13 & 0 & 0--2 &0--3   &1--4   & 0--2  & 0--2  & 9--26 \\
 & SFS+GB           & 0--2& 3--13 & 0--1 & 0--4 &0--2   &1--5  &  0--2 & 0--1  & 5--24 \\
 & SFS+KNN          & 1 & 3--15 & 0--1 & 0--4 & 0--2  & 1--2  & 0  & 0--1  & 7--23 \\
 & SFS+LR           & 0--1 & 5--13 & 0 & 0 &  0--2 &  1--3 &  0--2 &  0--1 & 7--18 \\
 & SFS+MLP          & 0--3& 3--16 & 0--1 & 0--3 &0--2   &1--4  &  0--4 & 0--1  & 7--26 \\
 & SFS+RF         & 1--3& 8--13 & 0 & 0--2 &0--2   &1--3  &  0--3 & 0--1  & 17--30 \\
 & SFS+SVM    & 1--3 & 5--15 & 0--1 & 0--2 & 1--2 & 1--4 & 0--2 & 0--3 & 10--22 \\
 & SFS+XGB          & 1--2 & 4--14 & 0 & 0--2 &0--2   &1--3   & 0--3  & 0--2  & 7--24 \\
\hline

 AdaBoost  & \multirow{6}{*}{RFECV}      & 1--3 & 10--18 & 1--4 & 0--1 & 2--3  &  4--5 & 1--2  & 0--2  & 21--39 \\
 DT      &         & 0--4 & 6--39 & 0--5 & 0--6 & 2--4   & 1--7   & 0--4   & 0--2  & 14--71 \\
 GB       &        & 1--4 & 21--39 & 1--5 & 1--6 & 3--4 & 5--7  & 1--4 & 0--2 & 33--71 \\

 LR      &   & 1--4 & 23--39 & 0--5 & 0--6 & 3--4  & 6--7   & 0--4   & 0--2 & 44--70 \\

 RF &  & 3--4 & 17--39 & 4--5 & 1--6 & 2--4  & 6--7  & 1--4  & 0--2 & 34--70 \\

XGB &     & 1--4 & 11--26 & 0--4 & 0--5 & 2--4  & 4--7   & 1--4  & 0--2  & 20--56 \\
\hline

\multicolumn{2}{l}{Lasso} & 2--4 & 26--35 & 0--3 & 4--5 & 3--4  & 6--7 & 4 & 12 & 49--62 \\
\hline

\multicolumn{2}{l}{Mann-Whitney U Test} & 4 & 30--34 & 5 & 6 & 3 & 7 & 3 & 2 & 60--64 \\
\hline
\end{tabular} }
\caption*{\footnotesize  --Values are presented as ranges (minimum--maximum) representing the number of features selected across all cross-validation folds.}
\end{table}

A clear and consistent pattern across all feature selection strategies is the dominance of MFCC features, which exhibit the widest and most stable selection ranges regardless of the method. In almost all configurations, particularly RFECV, LASSO, and Mann-Whitney U, MFCCs constitute the largest proportion of the selected subsets (often exceeding 50\% of total features). This consistency indicates that spectral envelope information is the most reliable and discriminative representation for capturing PD-induced vocal impairments. Importantly, even when aggressive dimensionality reduction is applied (e.g., SFS-based methods), MFCCs are rarely excluded, reinforcing their robustness across folds and classifiers.

Beyond MFCCs, several feature groups demonstrate consistent but secondary importance. Notably, intensity and harmonicity features are repeatedly selected across nearly all methods, particularly in RFECV and LASSO, where they appear in relatively narrow and stable ranges (e.g., intensity: 4--7, harmonicity: 2--4). This suggests that energy-related and voice quality characteristics provide complementary and reliable cues for PD detection, likely reflecting reduced vocal strength and irregular phonation patterns.

In contrast, jitter and shimmer features exhibit moderate but method-dependent variability. While they are frequently selected in RFECV and statistical methods (e.g., Mann-Whitney U consistently selects all jitter and shimmer features), their inclusion is less stable in Relief-F + SFS configurations. This indicates that although perturbation-based features are informative, their contribution is often context-dependent and influenced by feature interactions, rather than being independently dominant. A particularly important observation is the behavior of formant features, which show high stability despite their small dimensionality. Across multiple methods, including RFECV, LASSO, and even several SFS configurations, formants are consistently retained, often with minimal variation (e.g., 1--4 or full selection). This suggests that articulatory information captured by formants provides unique and non-redundant signals, complementing both spectral (MFCC) and perturbation features. Their stability is especially notable given that many prior studies underutilize them. Similarly, pulse features demonstrate moderate consistency, particularly in RFECV and LASSO, where at least 2–4 features are frequently retained. This indicates that temporal micro-variations in the speech signal, often overlooked, carry meaningful information related to PD-induced motor irregularities.

On the other hand, pitch features are the least consistently selected, typically appearing in very small ranges (0--2) across most methods, except for the Mann-Whitney U test, where both features are always selected. This suggests that pitch alone is not a strong standalone discriminator, but may still provide auxiliary prosodic information when combined with other features. From a methodological perspective, the differences between selection strategies reveal important insights. Relief-F + SFS combinations produce the most variable feature subsets, both in size and composition. This variability reflects the greedy and model-dependent nature of SFS, where feature inclusion is driven by incremental performance gains, making it sensitive to fold-level data fluctuations. Within this group, models such as SFS+RF, SFS+SVM, and SFS+MLP tend to select relatively richer subsets, suggesting their ability to leverage higher-order feature interactions. In contrast, RFECV exhibits the widest selection range, from highly compact subsets to nearly the full feature space (e.g., DT and GB reaching up to 71 features). This highlights its adaptive pruning mechanism, where feature elimination is guided by cross-validated performance. Despite this variability, RFECV consistently retains core feature groups, particularly MFCCs and intensity, indicating their fundamental importance.

LASSO demonstrates the highest stability among all methods, producing moderately compact subsets (49–62 features) with balanced representation across categories. Its embedded regularization effectively removes redundant features while preserving the most informative ones, resulting in low variance across folds.

Finally, the Mann–Whitney U test consistently selects a large and nearly fixed set of features (60–64), including all jitter, shimmer, and pitch features. While this confirms strong statistical differences between PD and control groups, it also highlights a limitation: univariate methods fail to account for inter-feature redundancy and interaction effects, leading to overly large subsets.

Overall, these results reveal that feature importance in PD detection is not uniform but hierarchical. MFCCs form the core representation, while intensity, harmonicity, and formants provide stable complementary information. Perturbation features (jitter, shimmer) contribute in a more context-dependent manner, and pitch plays a minor supporting role. The findings also demonstrate that reporting feature selection ranges across folds provides a more realistic measure of feature robustness, and that hybrid approaches (e.g., Relief-F + SFS) are effective in capturing diverse feature interactions, whereas RFECV and LASSO prioritize stability and generalization.

Crucially, the analysis moves beyond conventional MFCC-centric approaches by showing that a multi-dimensional combination of spectral, articulatory, and temporal features is essential for building a reliable and generalizable PD detection system.

\subsection{Hyperparameter tuning}
To ensure robust model selection and fair comparison across classifiers, we conducted hyperparameter tuning for all nine machine learning algorithms within the inner loop of a nested cross-validation framework. We adopted RandomizedSearch, a probabilistic search strategy that samples from specified distributions of hyperparameter values rather than exhaustively evaluating every possible combination. This approach strikes a balance between computational efficiency and search space coverage, especially when dealing with high-dimensional parameter spaces.

Table~\ref{tab:hyperparameter} summarizes the hyperparameter search ranges for each model. The search was guided by 5-fold cross-validation to ensure that selected hyperparameters generalized well across data splits. Performance was evaluated using the mean cross-validated score to identify the best configuration. This systematic tuning process ensured that each classifier operated under its most favorable configuration, enabling a fair and competitive experimental evaluation of PD detection using our proposed framework.
\begin{table}[!pt]
\centering
\caption{The tuned hyperparameters and the search range.
\label{tab:hyperparameter}}
\renewcommand{\arraystretch}{0.68}
\begin{tabular}{llp{.05cm}l}
\hline
Model & Hyperparameter & &Search range \\
\hline
\hline

\multirow{4}{*}{Logistic Regression (LR)}
    & Penalty & & l1, l2, elasticnet \\
    & C & & $10^{-4}$ to $10^4$ (logspace) \\
    & Solver & & liblinear, saga \\
    & Max Iterations & & 100, 200, 500, 1000 \\

\hline
\multirow{6}{*}{Random Forest (RF)}
    & n\_estimators & & 100, 300, 500 \\
    & Max Depth & & None, 10, 20, 50 \\
    & Min Samples Split & & 2, 5, 10 \\
    & Min Samples Leaf & & 1, 2, 4 \\
    & Max Features & & sqrt, log2, None \\
    & Bootstrap & & True, False \\

\hline
\multirow{4}{*}{SVM}
    & C & & 0.1, 1, 10, 100 \\
    & Kernel & & linear, rbf, poly \\
    & Degree & & 2, 3, 4 \\
    & Gamma & & scale, auto, 0.001, 0.01, 0.1, 1 \\

\hline
\multirow{4}{*}{KNN}
    & n\_neighbors & & 3, 5, 7, 9, 11 \\
    & Weights & & uniform, distance \\
    & Algorithm & & auto, ball\_tree, kd\_tree, brute \\
    & p & & 1, 2 \\

\hline
\multirow{6}{*}{Gradient Boosting (GB)}
    & n\_estimators & & 100, 200, 300 \\
    & Learning Rate & & 0.01, 0.05, 0.1 \\
    & Max Depth & & 3, 5, 7 \\
    & Subsample & & 0.7, 0.85, 1.0 \\
    & Min Samples Split & & 2, 5, 10 \\
    & Min Samples Leaf & & 1, 3, 5 \\

\hline
\multirow{6}{*}{Decision Tree (DT)}
    & Criterion & & gini, entropy \\
    & Max Depth & & None, 5, 10, 20 \\
    & Min Samples Split & & 2, 5, 10 \\
    & Min Samples Leaf & & 1, 2, 4 \\
    & Max Features & & sqrt, log2, None \\
    & ccp\_alpha & & 0.0, 0.01, 0.1 \\

\hline
\multirow{5}{*}{MLP}
    & Hidden Layer Sizes & & (100,), (50,50), (64,32) \\
    & Activation & & relu, tanh \\
    & Solver & & adam, sgd \\
    & Alpha & & 0.0001, 0.001, 0.01 \\
    & Learning Rate & & constant, adaptive \\

\hline
\multirow{9}{*}{XGBoost}
    & n\_estimators & & 100, 300, 500, 1000 \\
    & Max Depth & & 3, 6, 10 \\
    & Learning Rate & & 0.01, 0.1 \\
    & Subsample & & 0.7, 1.0 \\
    & Colsample Bytree & & 0.7, 1.0 \\
    & Gamma & & 0, 0.1 \\
    & Min Child Weight & & 1, 5 \\
    & reg\_alpha & & 0, 0.1 \\
    & reg\_lambda & & 1, 10 \\

\hline
\multirow{3}{*}{AdaBoost}
    & n\_estimators & & 50, 100, 200 \\
    & Learning Rate & & 0.01, 0.1, 0.5, 1.0 \\
    & Estimator & & DT(max\_depth=1/2), RF(n=100/300) \\

\hline

\hline
\end{tabular}%
\end{table}

\subsection{Comparing machine learning models}
To rigorously evaluate the effectiveness of different classification algorithms for PD detection from voice features, we conducted a comprehensive comparative study involving nine ML classifiers: LR, DT, RF, SVM, GB, XGBoost, AdaBoost, KNN, and MLP. Each classifier was assessed under four distinct feature selection paradigms, namely LASSO, Relief-F, Mann-Whitney U Test, and RFECV, combined with classifier-specific estimators.

Model performance was evaluated using stratified 10-fold cross-validation to ensure robustness and generalizability. A diverse set of evaluation metrics was considered, including accuracy, AUC-ROC, F1-score, precision, sensitivity, specificity, and Brier loss. All reported results represent the mean performance across folds along with their corresponding 95\% confidence intervals (CIs), thereby providing a statistically reliable comparison. Table~\ref{tab:performance-comparison} summarizes these results.
\begin{table}[!hpt]
\centering
\caption{Model performance comparison across different machine learning classifiers and feature selection methods. The best result for each evaluation metric is highlighted in bold. \label{tab:performance-comparison}}
\renewcommand{\arraystretch}{1}
\resizebox{1\textwidth}{!}{
\begin{tabular}{llccccccc}
\hline
Model &  \multirow{2}{*}{\makecell[t l]{Feature\\selection}} & \multicolumn{7}{c}{Performance with 95\% bootstrap confidence interval (CI)}\\
\cline{3-9}
& & Accuracy (\%) & AUC-ROC & F1 Score (\%) & Precision (\%) & Sensitivity (\%) & Specificity (\%) & Brier loss ($\downarrow$)\\
\hline
\hline

\multirow{4}{*}{LR}
 & LASSO & $79.78_{\text{76.9--82.2}}$ & $0.870_{\text{0.85--0.89}}$ & $79.45_{\text{76.5--82.1}}$ & $79.96_{\text{77.1--82.3}}$ & $78.00_{\text{73.7--81.3}}$ & $81.55_{\text{77.3--84.7}}$ & $0.145_{\text{0.13--0.16}}$\\
 & Relief-F+SFS(MLP) & $75.67_{\text{72.7--78.2}}$ & $0.838_{\text{0.81--0.86}}$ & $74.86_{\text{71.8--77.8}}$ & $76.02_{\text{72.8--78.4}}$ & $72.67_{\text{68.0--76.4}}$ & $78.67_{\text{74.6--82.0}}$ & $0.165_{\text{0.15--0.18}}$ \\
 & Mann-Whitney & $77.44_{\text{74.4--80.0}}$ & $0.860_{\text{0.83--0.88}}$ & $76.81_{\text{73.7--79.6}}$ & $77.81_{\text{74.6--80.1}}$ & $74.67_{\text{70.2--78.2}}$ & $80.22_{\text{76.0--83.6}}$ & $0.153_{\text{0.14--0.17}}$ \\
 & RFECV+RF & $79.67_{\text{76.8--82.0}}$ & $0.865_{\text{0.84--0.89}}$ & $79.07_{\text{76.1--81.7}}$ & $80.03_{\text{76.9--82.1}}$ & $76.89_{\text{72.4--80.2}}$ & $82.44_{\text{78.4--85.6}}$ & $0.149_{\text{0.14--0.16}}$ \\
\hline

\multirow{4}{*}{DT}
 & LASSO & $84.67_{\text{82.1--86.8}}$ & $0.835_{\text{0.81--0.86}}$ & $84.09_{\text{81.5--86.6}}$ & $84.86_{\text{82.3--87.0}}$ & $81.33_{\text{77.3--84.4}}$ & $88.00_{\text{84.4--90.7}}$ & $0.150_{\text{0.13--0.18}}$ \\
 & Relief-F+SFS(RF) & $85.89_{\text{83.4--87.9}}$ & $0.866_{\text{0.84--0.88}}$ & $84.96_{\text{82.5--87.5}}$ & $86.65_{\text{83.9--88.3}}$ & $80.89_{\text{76.7--84.2}}$ & $90.89_{\text{87.8--93.1}}$ & $0.138_{\text{0.12--0.16}}$ \\
 & Mann-Whitney & $84.22_{\text{81.6--86.3}}$ & $0.839_{\text{0.81--0.86}}$ & $83.41_{\text{80.7--85.9}}$ & $84.64_{\text{82.0--86.7}}$ & $79.56_{\text{75.3--82.9}}$ & $88.89_{\text{85.6--91.6}}$ & $0.155_{\text{0.13--0.18}}$ \\
 & RFECV+DT & $86.22_{\text{83.7--88.2}}$ & $0.861_{\text{0.84--0.89}}$ & $85.77_{\text{83.3--88.1}}$ & $86.50_{\text{83.9--88.5}}$ & $83.33_{\text{79.3--86.2}}$ & $89.11_{\text{85.6--91.6}}$ & $0.133_{\text{0.11--0.16}}$ \\
\hline

\multirow{4}{*}{RF}
 & LASSO & $94.89_{\text{93.2--96.1}}$ & $0.989_{\text{0.98--0.99}}$ & $94.86_{\text{93.3--96.2}}$ & $94.95_{\text{93.3--96.1}}$ & $94.44_{\text{91.8--96.2}}$ & $95.33_{\text{92.9--96.9}}$ & $0.059_{\text{0.05--0.07}}$ \\
 & Relief-F+SFS(XGB) & $93.89_{\text{92.1--95.2}}$ & $0.978_{\text{0.97--0.99}}$ & $93.82_{\text{92.1--95.3}}$ & $93.95_{\text{92.2--95.3}}$ & $93.11_{\text{90.2--95.1}}$ & $94.67_{\text{92.0--96.2}}$ & $0.061_{\text{0.05--0.07}}$ \\
 & Mann-Whitney & $\textbf{95.67}_{\text{94.0--96.8}}$ & $\textbf{0.990}_{\text{0.98--0.99}}$ & $\textbf{95.62}_{\text{94.1--96.9}}$ & $\textbf{95.72}_{\text{94.1--96.9}}$ & $95.11_{\text{92.4--96.7}}$ & $\textbf{96.22}_{\text{93.8--97.6}}$ & $0.056_{\text{0.05--0.06}}$ \\
 & RFECV+LR & $95.56_{\text{93.9--96.7}}$ & $\textbf{0.990}_{\text{0.98--0.99}}$ & $95.54_{\text{94.1--96.8}}$ & $95.61_{\text{93.9--96.7}}$ & $\textbf{95.56}_{\text{93.1--97.1}}$ & $95.56_{\text{93.1--97.1}}$ & $0.056_{\text{0.05--0.06}}$ \\
\hline

\multirow{4}{*}{SVM}
 & LASSO & $71.22_{\text{68.2--74.0}}$ & $0.774_{\text{0.74--0.80}}$ & $73.03_{\text{70.3--75.8}}$ & $71.76_{\text{68.6--74.5}}$ & $78.00_{\text{73.8--81.6}}$ & $64.44_{\text{59.8--68.7}}$ & $0.193_{\text{0.18--0.21}}$ \\
 & Relief-F+SFS(MLP) & $83.00_{\text{80.3--85.2}}$ & $0.880_{\text{0.87--0.91}}$ & $83.06_{\text{80.4--85.4}}$ & $83.40_{\text{80.3--85.3}}$ & $83.11_{\text{79.1--86.2}}$ & $82.89_{\text{78.9--86.0}}$ & $0.146_{\text{0.14--0.16}}$ \\
 & Mann-Whitney & $71.67_{\text{68.7--74.4}}$ & $0.788_{\text{0.76--0.81}}$ & $72.80_{\text{69.9--75.6}}$ & $72.09_{\text{68.8--74.7}}$ & $76.00_{\text{71.6--79.6}}$ & $67.33_{\text{62.7--71.3}}$ & $0.188_{\text{0.18--0.20}}$ \\
 & RFECV+DT & $74.56_{\text{71.6--77.2}}$ & $0.803_{\text{0.78--0.84}}$ & $75.32_{\text{72.3--77.9}}$ & $75.12_{\text{71.7--77.4}}$ & $77.11_{\text{72.9--80.7}}$ & $72.00_{\text{67.6--75.8}}$ & $0.177_{\text{0.17--0.19}}$ \\
\hline

\multirow{4}{*}{GB}
 & LASSO & $95.00_{\text{93.2--96.2}}$ & $0.986_{\text{0.98--0.99}}$ & $95.00_{\text{93.4--96.2}}$ & $95.05_{\text{93.3--96.2}}$ & $95.11_{\text{92.4--96.7}}$ & $94.89_{\text{92.2--96.4}}$ & $\textbf{0.041}_{\text{0.03--0.05}}$ \\
 & Relief-F+SFS(XGB) & $90.67_{\text{88.4--92.3}}$ & $0.960_{\text{0.95--0.97}}$ & $90.47_{\text{88.4--92.4}}$ & $90.85_{\text{88.6--92.4}}$ & $89.78_{\text{86.4--92.2}}$ & $91.56_{\text{88.4--93.8}}$ & $0.071_{\text{0.06--0.09}}$ \\
 & Mann-Whitney & $94.22_{\text{92.4--95.4}}$ & $0.983_{\text{0.98--0.99}}$ & $94.17_{\text{92.5--95.6}}$ & $94.30_{\text{92.6--95.6}}$ & $93.33_{\text{90.4--95.1}}$ & $95.11_{\text{92.4--96.7}}$ & $0.048_{\text{0.04--0.06}}$ \\
 & RFECV+LR & $95.56_{\text{93.9--96.7}}$ & $0.987_{\text{0.98--0.99}}$ & $95.54_{\text{94.0--96.8}}$ & $95.60_{\text{94.0--96.7}}$ & $95.33_{\text{92.7--96.9}}$ & $95.78_{\text{93.1--97.1}}$ & $\textbf{0.041}_{\text{0.03--0.05}}$ \\
\hline

\multirow{4}{*}{XGBoost}
 & LASSO & $93.33_{\text{91.4--94.7}}$ & $0.981_{\text{0.97--0.99}}$ & $93.36_{\text{91.5--94.8}}$ & $93.46_{\text{91.5--94.8}}$ & $93.56_{\text{90.4--95.3}}$ & $93.11_{\text{90.2--94.9}}$ & $0.050_{\text{0.04--0.06}}$ \\
 & Relief-F+SFS(XGB) & $92.78_{\text{90.8--94.2}}$ & $0.973_{\text{0.96--0.98}}$ & $92.62_{\text{90.7--94.3}}$ & $92.92_{\text{91.0--94.4}}$ & $91.33_{\text{88.2--93.6}}$ & $94.22_{\text{91.6--96.0}}$ & $0.057_{\text{0.05--0.07}}$ \\
 & Mann-Whitney & $93.00_{\text{91.1--94.4}}$ & $0.977_{\text{0.97--0.99}}$ & $93.00_{\text{91.2--94.6}}$ & $93.09_{\text{91.1--94.5}}$ & $93.11_{\text{90.2--95.1}}$ & $92.89_{\text{90.0--94.9}}$ & $0.054_{\text{0.05--0.07}}$ \\
 & RFECV+LR & $94.11_{\text{92.3--95.3}}$ & $0.982_{\text{0.98--0.99}}$ & $94.12_{\text{92.4--95.5}}$ & $94.16_{\text{92.3--95.5}}$ & $94.44_{\text{91.8--96.0}}$ & $93.78_{\text{91.1--95.6}}$ & $0.048_{\text{0.04--0.06}}$ \\
\hline

\multirow{4}{*}{AdaBoost}
 & LASSO & $89.44_{\text{87.2--91.2}}$ & $0.956_{\text{0.94--0.97}}$ & $89.70_{\text{87.6--91.5}}$ & $89.65_{\text{87.4--91.4}}$ & $91.56_{\text{88.4--93.8}}$ & $87.33_{\text{83.8--90.0}}$ & $0.190_{\text{0.19--0.19}}$ \\
 & Relief-F+SFS(AdaBoost) & $84.78_{\text{82.1--86.9}}$ & $0.922_{\text{0.90--0.94}}$ & $84.82_{\text{82.3--87.0}}$ & $84.90_{\text{82.2--86.9}}$ & $85.33_{\text{81.6--88.2}}$ & $84.22_{\text{80.4--87.1}}$ & $0.209_{\text{0.21--0.21}}$ \\
 & Mann-Whitney & $86.67_{\text{84.2--88.7}}$ & $0.944_{\text{0.93--0.96}}$ & $86.82_{\text{84.6--88.9}}$ & $86.86_{\text{84.4--88.8}}$ & $88.00_{\text{84.4--90.7}}$ & $85.33_{\text{81.6--88.2}}$ & $0.192_{\text{0.19--0.20}}$ \\
 & RFECV+LR & $89.67_{\text{87.4--91.4}}$ & $0.956_{\text{0.95--0.97}}$ & $89.84_{\text{87.8--91.7}}$ & $89.85_{\text{87.6--91.6}}$ & $91.56_{\text{88.2--93.8}}$ & $87.78_{\text{84.2--90.2}}$ & $0.191_{\text{0.19--0.19}}$ \\
\hline

\multirow{4}{*}{KNN}
 & LASSO & $83.22_{\text{80.7--85.4}}$ & $0.916_{\text{0.90--0.93}}$ & $81.75_{\text{78.9--84.5}}$ & $84.13_{\text{81.6--86.2}}$ & $75.56_{\text{71.3--79.1}}$ & $90.89_{\text{87.8--93.1}}$ & $0.119_{\text{0.10--0.14}}$ \\
 & Relief-F+SFS(MLP) & $86.89_{\text{84.4--88.8}}$ & $0.926_{\text{0.91--0.94}}$ & $86.42_{\text{84.0--88.7}}$ & $87.19_{\text{84.7--89.0}}$ & $84.00_{\text{80.2--86.9}}$ & $89.78_{\text{86.7--92.2}}$ & $0.100_{\text{0.09--0.11}}$ \\
 & Mann-Whitney & $84.00_{\text{81.4--86.1}}$ & $0.913_{\text{0.89--0.93}}$ & $82.52_{\text{79.7--85.2}}$ & $85.09_{\text{82.6--87.0}}$ & $75.78_{\text{71.6--79.3}}$ & $92.22_{\text{89.1--94.2}}$ & $0.119_{\text{0.10--0.14}}$ \\
 & RFECV+LR & $86.89_{\text{84.6--88.9}}$ & $0.946_{\text{0.93--0.96}}$ & $85.83_{\text{83.2--88.3}}$ & $87.65_{\text{85.5--89.5}}$ & $79.78_{\text{75.8--83.1}}$ & $94.00_{\text{91.1--95.8}}$ & $0.092_{\text{0.08--0.11}}$ \\
\hline

\multirow{4}{*}{MLP}
 & LASSO & $59.33_{\text{56.1--62.3}}$ & $0.740_{\text{0.60--0.68}}$ & $47.63_{\text{50.9--58.8}}$ & $66.05_{\text{56.4--62.9}}$ & $49.56_{\text{44.7--54.0}}$ & $69.11_{\text{64.4--73.1}}$ & $0.375_{\text{0.35--0.40}}$ \\
 & Relief-F+SFS(MLP) & $85.89_{\text{83.4--88.0}}$ & $0.926_{\text{0.91--0.94}}$ & $85.80_{\text{83.4--88.1}}$ & $86.19_{\text{83.6--88.1}}$ & $85.11_{\text{81.3--88.0}}$ & $86.67_{\text{83.1--89.3}}$ & $0.105_{\text{0.09--0.12}}$ \\
 & Mann-Whitney & $66.00_{\text{62.7--68.9}}$ & $0.792_{\text{0.67--0.74}}$ & $55.73_{\text{59.3--66.7}}$ & $72.38_{\text{63.2--69.4}}$ & $58.22_{\text{53.3--62.4}}$ & $73.78_{\text{69.3--77.3}}$ & $0.276_{\text{0.25--0.30}}$ \\
 & RFECV+LR & $72.00_{\text{68.9--74.7}}$ & $0.828_{\text{0.75--0.81}}$ & $68.14_{\text{66.8--73.4}}$ & $75.50_{\text{69.3--75.1}}$ & $66.00_{\text{61.3--70.0}}$ & $78.00_{\text{73.8--81.3}}$ & $0.202_{\text{0.19--0.22}}$ \\

\hline

\end{tabular}
}
\end{table}

Among all approaches, ensemble methods consistently achieve superior performance. Random Forest attains the best overall results under the Mann-Whitney feature selection scheme, achieving 95.67\% accuracy (95\% CI 94.0\%–96.8\%), 0.990 AUC-ROC (95\% CI 0.98–0.99), and 95.62\% F1-score (95\% CI 94.1\%–96.9\%), along with the highest specificity 96.22\% (95\% CI 93.5\%–97.6\%). Gradient Boosting closely follows, reaching 95.56\% accuracy (95\% CI 93.9\%–96.7\%) under RFECV+LR and yielding the lowest Brier loss 0.041 (95\% CI 0.03–0.05), indicating excellent probability calibration. XGBoost also demonstrates stable and competitive performance across all configurations, with accuracy consistently above 93\%. These findings highlight the effectiveness of ensemble learners in capturing complex, nonlinear relationships inherent in voice-based Parkinson’s disease data.

 Wrapper-based RFECV produces consistently strong and stable results across multiple classifiers, as reflected by narrow confidence intervals. However, the Mann–Whitney method also achieves top-tier performance when paired with RF, suggesting that retaining a larger set of statistically significant features can be advantageous for ensemble models. Notably, the hybrid Relief-F+SFS approach significantly enhances performance for models that are sensitive to feature representation, such as SVM and MLP. For example, SVM improves from 71.22\% (95\% CI 68.2\%–74.0\%) (LASSO) to 83\% (95\% CI 80.3\%–85.2\%), while MLP increases from 59.33\% (95\% CI 56.1\%–62.3\%) to 85.89\% (95\% CI 83.4\%–88.0\%), indicating that sequential refinement of feature subsets effectively improves separability in the feature space.

In contrast, simpler models exhibit comparatively moderate and less stable performance. LR remains consistent across feature selection methods, peaking at 79.78\% accuracy (95\% CI 76.9\%–82.2\%), while DT benefits from RFECV, achieving 86.22\% (95\% CI 83.7\%–88.2\%). KNN demonstrates clear dependence on feature quality, improving to 86.89\% (95\% CI 84.6\%–88.9\%) under both Relief-F+SFS and RFECV+LR.
From a calibration perspective, ensemble models again outperform others, achieving the lowest Brier scores (e.g., 0.041 for GB and 0.056 for RF), which is critical for reliable clinical decision-making. Conversely, higher Brier scores observed for SVM and MLP in certain configurations indicate less reliable probabilistic predictions.

Overall, the results indicate that the combination of robust feature selection and ensemble learning, particularly RF with Mann-Whitney and GB with RFECV or LASSO, provides the most effective balance between accuracy, stability, and calibration, making these approaches highly suitable for practical voice-based Parkinson’s disease detection systems.

\subsection{Comparing with existing approaches}
We compared the performance of our proposed framework with three state-of-the-art recent models that leverage various deep learning approaches for audio-based PD detection. These models serve as baselines for comparison with our approach. To ensure a fair and consistent comparison, we used the same set of 71 handcrafted audio features, extracted from our Bengali conversational speech dataset, as input for each of the baseline models.

\begin{itemize}
    \item {EF-LSTM (Early Fusion with LSTM) \citep{cai2025multimodal}:} In this approach, features processed through a temporal convolution are concatenated and passed into a single LSTM network. Specifically, a temporal convolution layer with 1 layer, 64 filters, a kernel size of 3, ReLU activation, and a dropout rate of 0.2 is first applied to capture local temporal dependencies. The resulting features are then fed into a bidirectional LSTM layer consisting of 1 layer with 128 hidden units and a dropout rate of 0.3. The hidden state from the LSTM serves as the sequence encoding and is passed through a fully connected layer (64 units, ReLU activation, dropout 0.5) followed by a classification layer to produce the final prediction.

    \item {LF-LSTM (Late Fusion with LSTM) \citep{cai2025multimodal}:} This method processes each feature set independently through separate streams. Unlike EF-LSTM, no convolutional layers are applied prior to the LSTM. Instead, each feature set is directly fed into an individual LSTM network consisting of 1 layer with 128 hidden units and a dropout rate of 0.3. The hidden states from each feature-specific LSTM are then concatenated and passed into a final LSTM layer, whose hidden representation is used to generate the final classification output.

    \item {1D-CNN \citep{9426437}:} The proposed model utilizes a 1-dimensional convolutional neural network (1D-CNN) combined with the multilayer perceptron (MLP) for classification. The architecture comprises three convolutional layers: the first with 16 filters and a kernel size of 64, the second with 32 filters and a kernel size of 32, and the third with 64 filters and a kernel size of 16. Each convolutional layer employs ReLU activation and is followed by a max-pooling layer with a pool size of 2 and a stride of 2. The convolutional stack is followed by an MLP with 128 hidden units and ReLU activation. A fully connected output layer with a sigmoid activation function is used for final binary classification. This 1D CNN+MLP model includes approximately 60,000 trainable parameters and stands as an outperformer model for the study \citep{9426437} using the PC GITA dataset \citep{orozco2014new}.
\end{itemize}

As shown in Table~\ref{tab:baseline-model}, our framework significantly outperforms all three baseline approaches across all evaluation metrics. Specifically, our model achieves the highest classification accuracy of 95.67\% (95\% CI 94.0\%–96.8\%), compared to 94.77\% (95\% CI 93.1\%–96.0\%) for EF-LSTM, 92.56\% (95\% CI 90.5\%–94.0\%) for LF-LSTM, and 94.11\% (95\% CI 92.3\%–95.4\%)for 1D-CNN. Similarly, our framework yields the best F1 score 95.62\% (95\% CI 94.1\%–96.9\%) and AUC-ROC 0.990 (95\% CI 0.98–0.99), indicating superior predictive power and discrimination capability across class boundaries. A notable observation is that our framework consistently reports narrower confidence intervals across all metrics compared to the baseline models. This indicates greater stability and lower variance in performance across different cross-validation folds, suggesting that the model generalizes more reliably to unseen data.

\begin{table}[!ht]
    \centering
     \caption{Comparison of classification performance between the proposed framework and state-of-the-art deep learning approaches on the BenSParX dataset.}
    \begin{tabular}{lccc}
    \hline
         Approach & Accuracy (\%) &F1 Score  (\%)& AUC-ROC \\
         \hline\hline

         LF-LSTM~\citep{cai2025multimodal} &$92.56_{\text{90.5--94.0}}$ &$92.37_{\text{90.4-94.0}}$&$0.970_{\text{0.95--0.97}}$ \\
        EF-LSTM~\citep{cai2025multimodal} &$94.77_{\text{93.1--96.0}}$ &$94.51_{\text{92.8-95.9}}$&$0.983_{\text{0.97--0.98}}$\\
         1D-CNN~\citep{9426437} &$94.11_{\text{92.3--95.4}}$ &$93.86_{\text{92.1-95.4}}$&$0.985_{\text{0.97--0.99}}$\\ \hline
         This study (RF with Mann-Whitney U)& \textbf{$95.67_{\text{94.0--96.8}}$} &\textbf {$95.62_{\text{94.1--96.9}}$} &\textbf{$0.990_{\text{0.98--0.99}}$}         \\

         \hline

         \hline
    \end{tabular}
    \label{tab:baseline-model}
\end{table}

\subsection{External validation}
To further demonstrate the proposed framework's applicability, we apply it to state-of-the-art audio datasets for detecting PD and compare its performance with the existing studies on the same datasets.
Table \ref{tab:Existing works} presents the comparison that is based on three key evaluation metrics: accuracy, F1 score, and AUC.

\begin{table}[!ht]
\centering
    \caption{Performance comparison of our proposed framework on other datasets for detecting PD.} \label{tab:Existing works}
\begin{tabular}{llccc}
    \hline
   Dataset & Study &Accuracy (\%) &F1 score (\%) & AUC-ROC \\
    \hline\hline

    \multirow{9}{*}{Sakar18 \citep{parkinsons_disease_classification_470}}
    & \citet{el2022enhancing} & 90 & ---&--- \\
     & \citet{sakar2019comparative} & 86 & ---&--- \\
      & \citet{gunduz2019deep} & 86.9 & ---&--- \\
       & \citet{gunduz2019parkinson} & 88.1 & ---&--- \\
        & \citet{gunduz2021efficient} & 91.2 & ---&--- \\
         & \citet{jeong2024machine} & 92 & ---&--- \\

    &\citet{10.1145/3441417.3441425} & 92.2& ---&--- \\
    &\citet{8741725}& 94.89 & ---&--- \\
    \cmidrule(r){2-5}
     &This study (RF with Mann-Whitney U) & \textbf{95.35 }& \textbf{95.3}&\textbf{0.986} \\

    \hline
    \multirow{5}{*}{Little \citep{parkinsons_174}}
    &\citet{7042180} &80 & ---&---\\
    & \citet{balaha2025comprehensive} &95.67 &90.57&--- \\
    & \citet{Hadjaidji_2025} & 97.44 & ---&--- \\
    & \citet{ecsa-11-20481} & 95& 97&0.980 \\
    \cmidrule(r){2-5}
    &This study (RF with Relief-F+SFS)& \textbf{98.66}&\textbf{98.63} &\textbf{0.997} \\

    \hline
    \multirow{4}{*}{Naranjo \citep{parkinson_dataset_with_replicated_acoustic_features__489}}
    & \citet{NARANJO2017147} & 86.2 & ---&--- \\
     & \citet{polat2020parkinson} & 82.1 & ---&--- \\
      & \citet{ghaheri2024diagnosis} & 85.42 & ---&--- \\
       \cmidrule(r){2-5}
      &This study (SVM with RFECV+GB)& \textbf{87.91}&\textbf{86.82} &\textbf{0.929} \\

     \hline
    \multirow{3}{*}{MDVR-KCL \citep{jaeger2019mobile}}
    & \citet{di2024machine} & 92.3 & ---&--- \\
     & \citet{balaha2025comprehensive} & 94.03 &85.67&--- \\
     \cmidrule(r){2-5}
     &This study (XGBoost with RFECV+GB) & \textbf{98.59 }& \textbf{99.19}&\textbf{0.996} \\
    \hline

  \hline
    \end{tabular}
\end{table}

For the most widely used Sakar18 dataset~\citep{parkinsons_disease_classification_470}, our proposed framework outperforms most existing methods, achieving a high accuracy of 95.35\% and an F1 score of 95.3\%. Notably, we also report an AUC of 0.986, which highlights the robustness of our model in distinguishing Parkinson's patients from healthy individuals by using the Mann-Whitney U test for feature selection and RF as the classifier. This marks a significant improvement over previous works, such as those by \citet{8741725} (94.89\% accuracy). For the Little dataset \citep{parkinsons_174}, our framework achieved outstanding performance by combining Relief-F and Sequential Forward Selection (SFS) for feature selection, followed by classification using Random Forest. This approach resulted in an accuracy of 98.66\%, an F1 score of 98.63\%, and an AUC of 0.997. These results clearly surpass those of other leading approaches, including \citet{Hadjaidji_2025} (97.44\% accuracy) and \citet{ecsa-11-20481} (95\% accuracy and 97\% F1 score), establishing our framework as the top performer on this dataset. On the Naranjo dataset, our proposed framework outperforms \citet{NARANJO2017147}, with an accuracy of 87.91\%, a matching F1 score of 86.82\%, and a strong AUC of 0.929. This performance was obtained using RFECV with Gradient Boosting for feature selection and SVM as the classifier. For the MDVR-KCL dataset, our framework achieves an accuracy of 98.59\%, an F1 score of 99.19\%, and an AUC of 0.996 by using RFECV with Gradient Boosting for feature selection and XGBoost as the classifier. This performance compares favorably with the work of \citet{di2024machine} (92.3\% accuracy) and \citet{balaha2025comprehensive} (94.03\% accuracy), further showcasing the effectiveness of our approach across diverse datasets.

Overall, our proposed framework consistently achieves superior results across all four datasets compared to the existing approaches, demonstrating its effectiveness for PD detection from voice recordings in different languages, modalities, and acoustic feature sets. The high accuracy, F1 score, and AUC values across multiple datasets highlight the robustness and generalizability of BenSParX, reinforcing its potential for real-world applications in healthcare.

\subsection{Explainability}

\begin{figure}[!ht]
    \centering
    \includegraphics[width=.7\textwidth]{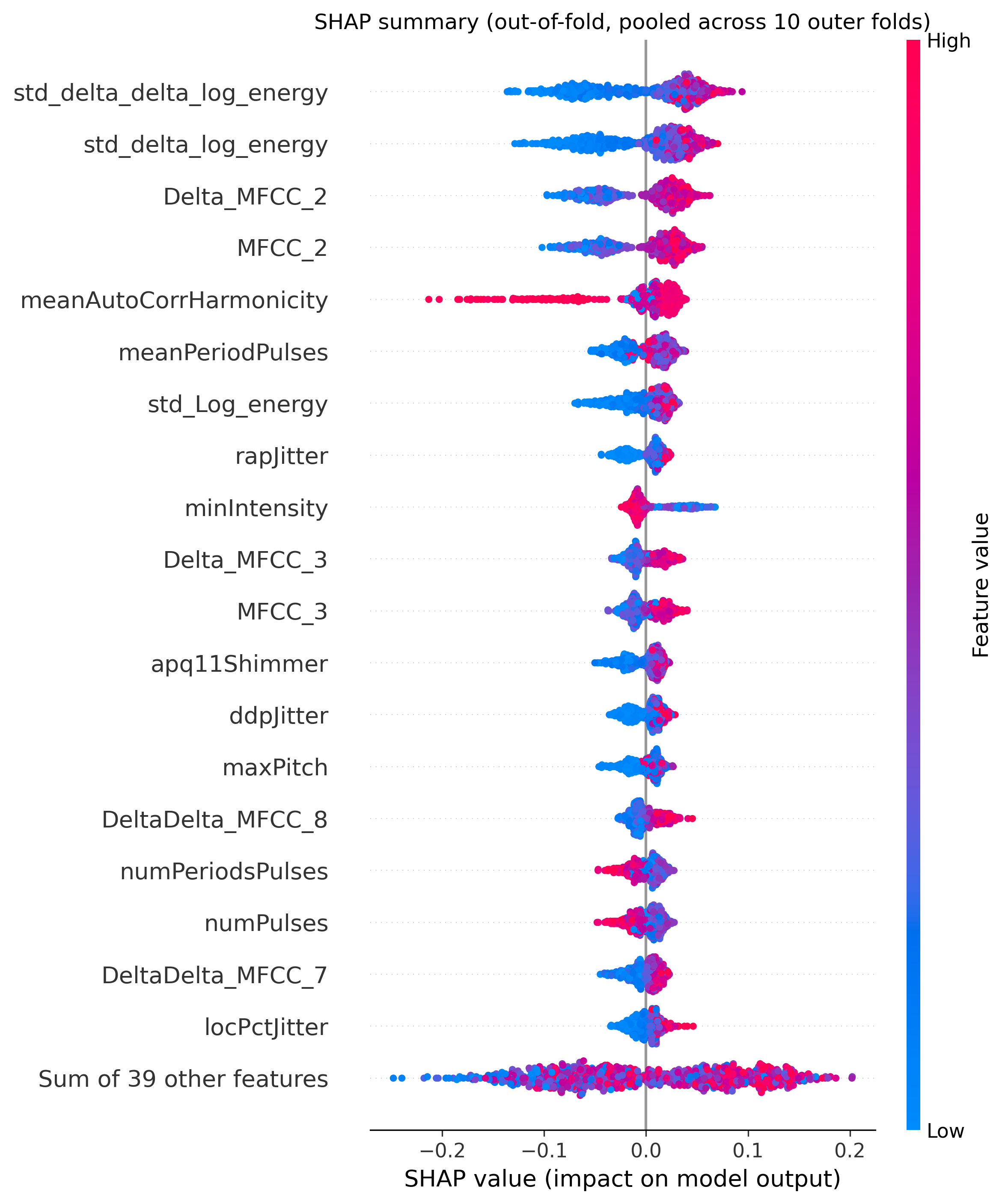}
    \caption{SHAP summary plot illustrating the impact of top features on the best-performing model's output for PD.}
    \label{fig:SHAP}
\end{figure}

To enhance model transparency and clinical relevance, we applied SHAP to interpret the contributions of acoustic features using the out-of-fold procedure described in Section \ref{sec:explan}. Across the 10 outer folds, the Mann-Whitney U test selected between 60 and 64 features per fold, with 58 features selected in every fold. Figure \ref{fig:SHAP} presents the SHAP summary plot for the top 20 of these 58 consistently selected features, pooled across all 10 outer folds' out-of-fold predictions, ranked along the Y-axis by mean SHAP value. The X-axis represents the SHAP value for each instance, indicating the direction and magnitude of each feature's contribution to the prediction. Positive SHAP values shift the prediction toward PD, while negative values shift it toward Healthy Control (HC). Each dot represents a single held-out instance, with color indicating the magnitude of the feature value, red for high and blue for low.

The most influential and consistently-ranked features were  \texttt{std\_delta\_delta\_log\_energy},\texttt{std\_delta\_log\_energy}, \texttt{Delta\_MFCC\_2}, \texttt{MFCC\_2}, and \texttt{meanAutoCorrHarmonicity}. These five features were selected in all 10 outer folds and their mean-SHAP rank varied by no more than 2–3 positions across folds (Supplementary material), indicating stable, fold-independent importance rather than an artifact of a single train/test split. These features reflect variability in vocal energy and harmonic structure, attributes often affected in PD due to hypophonia and impaired respiratory-laryngeal coordination. Increased variability in these features (as seen in red data points with positive SHAP values) pushed predictions toward PD, consistent with findings that vocal energy instability is a prominent early symptom of PD-related dysarthria~\citep{azevedo2003acoustic,favaro2024unveiling}. Several MFCC-related features (e.g., \texttt{MFCC\_2}, \texttt{Delta\_MFCC\_2}, \texttt{MFCC\_3}) also had high and stable impact, capturing changes in the speech spectrum consistent with articulatory imprecision and monoprosody – hallmarks of PD speech~\citep{upadhya2019discriminating,rios2024automatic}.

Perturbation features such as \texttt{rapJitter} and \texttt{apq11Shimmer}, and pulse-related features such as \texttt{mean\allowbreak Period\allowbreak Pulses}, were likewise selected in all 10 folds, though with somewhat wider rank ranges (Supplementary Table S1), reinforcing the model's reliance on frequency and amplitude micro-instabilities and voicing regularity, both well-documented consequences of impaired vocal fold control and bradykinesia/rigidity in PD~\citep{favaro2023multilingual,tsanas2011nonlinear}. Harmonicity features (\texttt{meanAutoCorrHarmonicity}) were also consistently influential, with lower harmonicity pushing predictions toward PD, indicating reduced periodicity and glottal insufficiency~\citep{little2009suitability,cao2025speech}.

\begin{figure}[!ht]
    \centering
    \includegraphics[width=.8\textwidth]{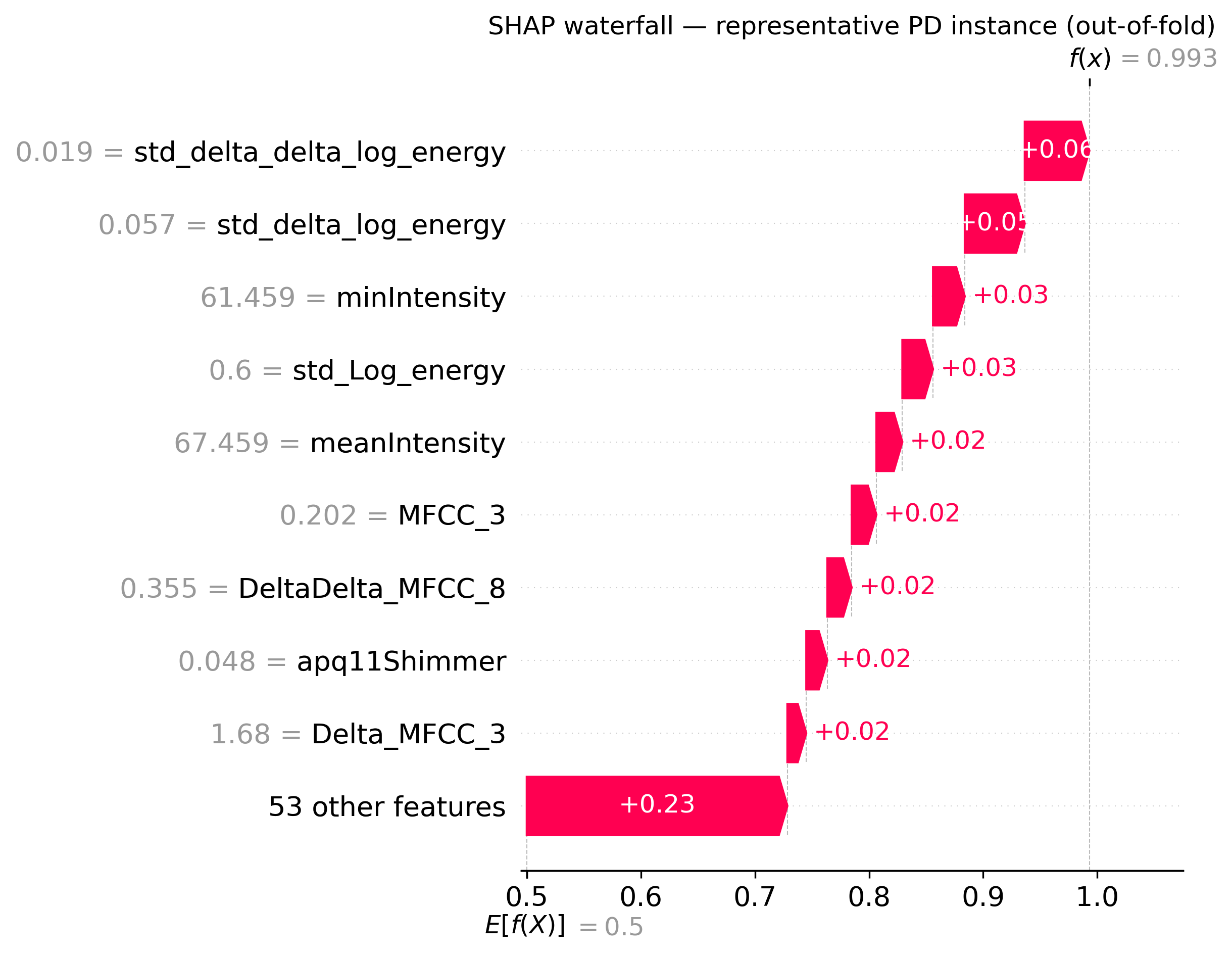}
    \caption{SHAP waterfall plot for a PD detection instance, illustrating the top 10 feature values contributing to the final model prediction.}
    \label{fig:WATERFALL-pd}
\end{figure}

Figure \ref{fig:WATERFALL-pd} illustrates the top-10 contributing features toward a representative PD instance, explained using the model from the specific outer fold in which that instance was held out. The baseline prediction begins at 0.5, the mean out-of-fold predicted probability, and shifts to a final probability of 0.993, indicating strong confidence in the PD classification. This confidence is driven by clinically meaningful acoustic markers consistent with known speech impairments in PD, including instability in energy (\texttt{std\_delta\_delta\_log\_energy}, \texttt{std\_delta\_log\_energy}, \texttt{std\_Log\_energy}) and reduced intensity (\texttt{minIntensity}, \texttt{meanIntensity}).

\begin{figure}[!ht]
    \centering
    \includegraphics[width=.8\textwidth]{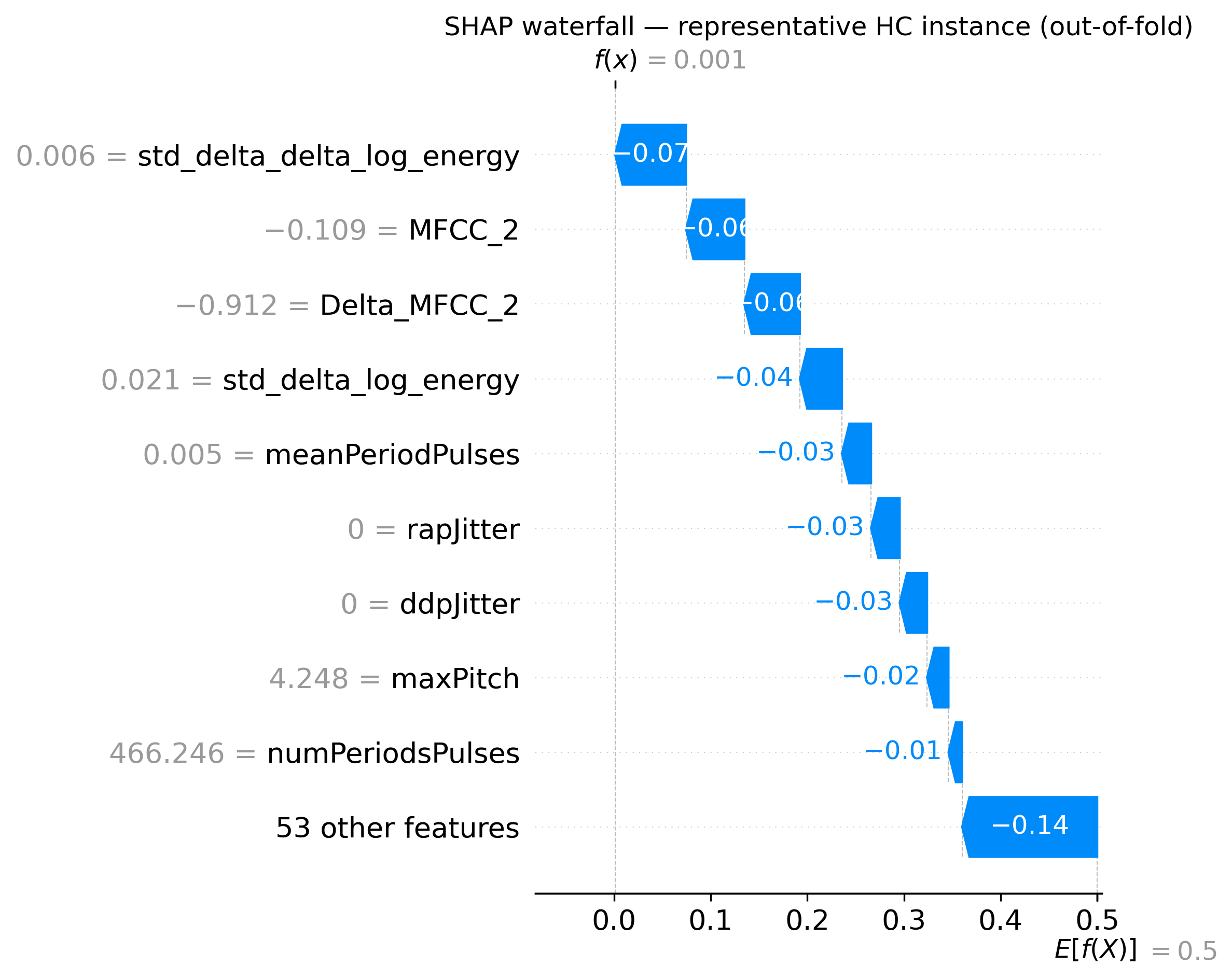}
    \caption{SHAP waterfall plot for a healthy control (HC) detection instance, illustrating the top 10 feature values contributing to the final model prediction.}
    \label{fig:WATERFALL-hc}
\end{figure}

Figure \ref{fig:WATERFALL-hc} illustrates the corresponding explanation for a representative HC instance, again using its own held-out fold's model. The prediction shifts from the 0.5 baseline down to 0.001, reflecting near-certain confidence in the HC classification. This downward shift is driven primarily by low values of \texttt{std\_delta\_delta\_log\_energy}, \texttt{MFCC\_2}, and \texttt{Delta\_MFCC\_2}, together with stable perturbation features (\texttt{rapJitter}, \texttt{ddpJitter} at or near zero), consistent with regular phonation and stable articulation typical of healthy speech.

\section{Discussion}
\subsection{Findings and observations}

Analysis of model performance and behavior yielded several observations regarding the proposed approach for detecting PD from Bengali conversational speech. Feature selection using the Mann-Whitney U test resulted in a subset of 60--64 features associated with the best-performing models under our experimental setup. This subset included formant- and pulse-related features, which have been less frequently reported in prior PD-focused speech studies. In our dataset, these features were consistently retained during selection and were associated with improved classification outcomes; however, we did not perform analyses to isolate their independent effects or to assess potential confounding.
The inclusion of pulse features may reflect sensitivity to aspects of voicing regularity and vocal effort, which are known to be affected in PD, while formant-related features may capture articulatory characteristics. In addition, a large proportion of MFCC features (30--34 out of 39), along with most jitter, shimmer, harmonicity, intensity, and pitch features, were also retained. Overall, these results suggest that a broad and diverse acoustic feature space may be useful for PD detection in this context, although further controlled studies are needed to confirm the specific contribution of individual feature groups.

Interestingly, performance across several classifiers did not vary significantly, suggesting that the extracted features were both robust and transferable across models. To evaluate our proposed framework's generalizability, we tested it on several benchmark PD voice datasets in different languages and compared it with existing studies. The results showed that the best-performing feature selection and classifier combinations varied across datasets. Feature selection techniques such as Mann-Whitney U, RFECV, Relief-F, LASSO, and SFS proved effective when paired with different classifiers, depending on the dataset. This variation underscores that no fixed combination works best universally. Our framework's ability to adapt to these variations highlights the need for a flexible and generic framework for reliable PD detection across diverse data sources.

Our framework outperformed recent deep learning models, despite using the same handcrafted features. Unlike the complex EF-LSTM, LF-LSTM, and 1D-CNN architectures, it delivered more consistent results with lower variability across validation folds, indicating stronger generalizability. This highlights the effectiveness of combining strategic feature selection with classical classifiers over deep models in certain settings. Overall, the proposed framework proves to be a more stable and practical solution for PD detection from speech.

\subsection{Limitations}
While this study demonstrates strong performance and clinical relevance, several limitations should be acknowledged.

\textit{First}, the study is based on a relatively modest sample size, which may limit statistical power and increase the risk of overfitting, particularly given the high dimensionality of the feature space. Although cross-validation was employed to mitigate this risk, larger and more diverse cohorts are necessary to confirm the robustness and generalizability of the findings.

\textit{Second}, the framework relies on a single short recording session (1 to 2 minutes) per participant. This limits the ability to capture temporal variability in speech, an important consideration for PD, where vocal symptoms may fluctuate throughout the day or in response to medication and fatigue. Although this design is consistent with several existing datasets (e.g., Sakar18~\citep{parkinsons_disease_classification_470}), it constrains dynamic assessment and intra-speaker variability analysis.

\textit{Third}, the dataset captures speech from individuals within 1.5 years of diagnosis, aligning with our goal of early-stage PD detection. However, limited clinical metadata (e.g., disease severity, medication state, and progression markers) prevent stratification of patients into clinically meaningful subtypes or prognostic groups. As a result, the model does not account for heterogeneity within PD and cannot support subtype-specific analysis or disease progression modeling.

\textit{Finally}, the dataset is drawn from a relatively homogeneous demographic, with limited variation in regional dialects, socio-economic background, or educational level. These factors, which can influence speech characteristics, were not explicitly controlled. This may affect the model's generalizability to broader, more linguistically and socially diverse populations, particularly across different Bengali-speaking regions.

\subsection{Future work}
Future research could extend the proposed framework in several directions.

\textit{First}, from the dataset perspective, incorporating multiple and longer recording sessions per participant, collected at different times and under varying medication states, would allow modeling of intra-speaker variability, which is particularly relevant for PD due to symptom fluctuations influenced by fatigue and medication cycles.

\textit{Second}, collecting longitudinal data with detailed clinical metadata such as disease severity, Unified Parkinson's Disease Rating Scale (UPDRS) scores, and medication history could enable the model to support PD staging and progression monitoring, expanding its clinical utility beyond early detection.

\textit{Third}, expanding the dataset to include participants from diverse regional, socio-economic, and dialectal backgrounds would improve the model's generalizability to the broader Bengali-speaking population and mitigate potential biases arising from demographic homogeneity.

\textit{Fourth}, future work could enrich the audio modality by incorporating a broader range of speech tasks, including sustained vowel phonations, isolated phonemes, and standardized sentence reading. These different vocalizations offer complementary acoustic and articulatory cues, which may enable deeper investigations such as phonation-type specific feature analysis, task-dependent symptom variability, and cross-language phonation modeling.
Fifth, integrating multimodal data sources, such as handwriting samples, gait recordings, or facial expressions, could lead to more comprehensive and accurate PD detection systems, especially in cases where vocal symptoms alone are insufficient for reliable diagnosis.

\textit{Finally}, from a methodological perspective, future work could explore alternative feature extraction techniques, particularly those involving the transformation of spectrograms or MFCC representations into image-like formats (e.g., \citep{iyer2023machine}) and the application of deep learning-based audio feature extractors such as VGGish~\citep{kurada2020poster} or other pretrained convolutional neural networks~\citep{van2024innovative}. Such approaches may facilitate more effective representation learning and enable transfer learning, thereby improving performance even in the presence of limited training data.

\section{Conclusion}
This study introduces BenSParX, a novel Bengali conversational speech dataset with a robust and explainable machine learning framework for PD detection, marking the first such effort in this linguistically underrepresented domain. Our contributions are two-fold : (i) we present the first publicly available Bengali PD speech dataset, collected in real-world settings through natural telephone conversations, and (ii) we propose a comprehensive and interpretable ML pipeline that integrates diverse acoustic feature categories, rigorous feature selection strategies, and state-of-the-art classifiers.
Through extensive experiments involving nine machine learning models and four feature selection techniques, we demonstrate that our framework consistently outperforms deep learning baselines on Bengali datasets and state-of-the-art approaches on existing datasets. SHAP-based explainability further validates that the most predictive features align with clinically established speech impairments in PD, including energy instability, reduced harmonicity, jitter, shimmer, and articulatory imprecision.

By combining clinical relevance, technical rigor, and model interpretability, this study sets a new benchmark for voice-based PD detection, particularly in low-resource languages. This work paves the way for more inclusive, accessible, and trustworthy AI-driven tools for early neurological screening and underscores the value of culturally and linguistically adaptive machine learning in digital health applications.

\section*{Data availability statement}
The labelled dataset created in this study is available at \url{https://github.com/riadEDU/BenSParX}.

\section*{Code availability statement}
The code in this study is available at \url{https://github.com/riadEDU/BenSParX}.
\section*{Declaration of competing interest}
The authors declare that they have no conflict of interest.

\section*{Ethics statement}
This study was conducted in accordance with the ethical principles outlined in the Declaration of Helsinki. Ethical approval was obtained from the Institutional Review Board (IRB) of Chittagong University of Engineering and Technology (approval number CSE201871). All participants provided informed consent prior to participation.

\section*{Declaration of Generative AI and AI-assisted technologies in the writing process}
During the preparation of this work, the authors used ChatGPT to improve their language and readability. After using this tool, the authors reviewed and edited the content as needed and took full responsibility for the content of the publication.

\section*{Funding}
This research received no external funding.

\section*{CRediT authorship contribution statement}
\textbf{Riad Hossain:} Conceptualization, Methodology, Formal analysis, Data Curation, Software, Writing -- Original Draft.
\textbf{Muhammad Ashad Kabir:} Conceptualization, Methodology, Formal analysis, Software, Writing -- Review \& Editing, Visualization, Validation.
\textbf{Arat Ibne Golam Mowla:} Data Curation, Visualization. \textbf{Animesh Chandra Roy:} Writing -- Original Draft, Supervision. \textbf{Ranjit Kumar Ghosh:} Data Curation, Validation.



\clearpage
\section*{Supplementary Material}
\addcontentsline{toc}{section}{Supplementary Material}

\setcounter{figure}{0}
\setcounter{table}{0}
\setcounter{equation}{0}
\setcounter{section}{0}

\renewcommand{\thefigure}{S\arabic{figure}}
\renewcommand{\thetable}{S\arabic{table}}
\renewcommand{\theequation}{S\arabic{equation}}
\renewcommand{\thesection}{S\arabic{section}}
\renewcommand{\thesubsection}{S\arabic{section}.\arabic{subsection}}

\section{Feature extraction}

\paragraph{Mel-Frequency Cepstral Coefficients (MFCCs)}

MFCCs are among the most widely used features in speech and audio signal processing due to their strong alignment with the characteristics of human auditory perception~\citep{di2024machine}. Derived from the mel-frequency cepstrum, MFCCs provide a compact representation of the short-term power spectrum of an audio signal by emphasizing frequency bands that are perceptually more relevant to human hearing.
Each MFCC represents a specific aspect of the signal’s spectral envelope. Typically, 12 to 20 coefficients are used to capture key spectral characteristics. In our study, we extracted 13 base MFCCs (coefficients 0 to 12), along with their first-order derivatives ($\Delta$ MFCCs) and second-order derivatives ($\Delta^2$ MFCCs), resulting in a total of 39 MFCC-based features per segment. This inclusion of dynamic features enhances the representation of temporal variations in speech, which are critical for capturing the instability present in Parkinsonian speech.

\begin{figure}[!ht]
    \centering
    \begin{subfigure}[b]{0.49\linewidth} 
        \centering
        \includegraphics[width=\textwidth]{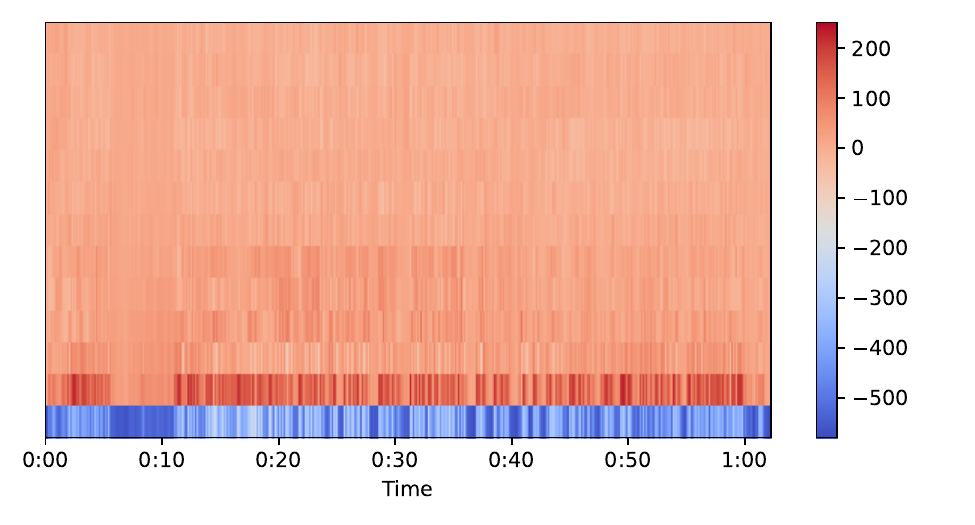}
        \caption{MFCC for a PD.}
        \label{fig:mfcc-pd}
    \end{subfigure}
    \begin{subfigure}[b]{0.49\linewidth}
        \centering
        \includegraphics[width=\textwidth]{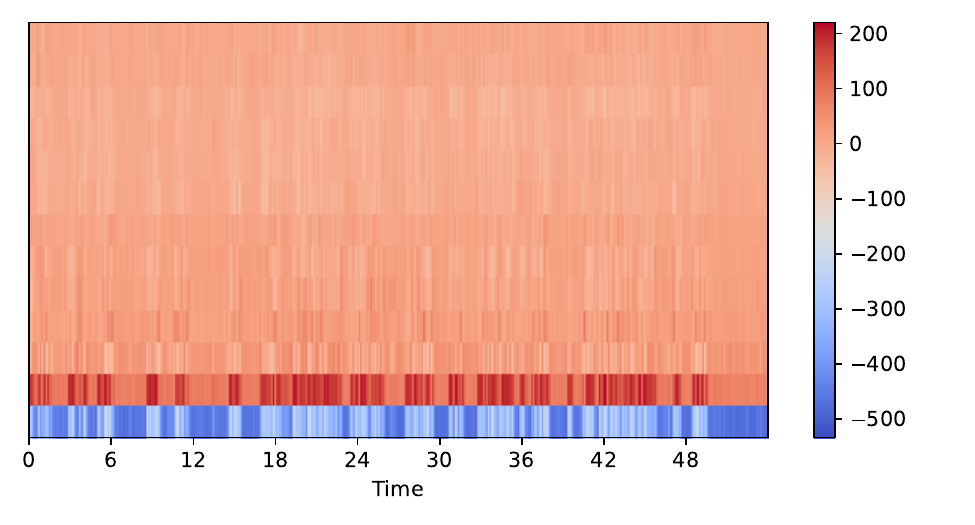}
        \caption{MFCC for an HC.}
        \label{fig:mfcc-hc}
    \end{subfigure}
    
    \caption{MFCC analysis for a PD and an HC. PD and HC correspond to randomly selected subjects from the dataset and are presented for illustrative purposes only. The plots are derived from the original raw audio recordings, which have varying durations; consequently, the time axes are not standardized across samples.}
    \label{fig:mfcc}
\end{figure}

Figure~\ref{fig:mfcc} presents a visual comparison of MFCC spectrograms extracted from a PD patient and an HC. The PD spectrogram (Figure~\ref{fig:mfcc-pd}) exhibits greater spectral irregularities and reduced contrast across time, particularly in lower frequency regions. In contrast, the HC spectrogram (Figure~\ref{fig:mfcc-hc}) shows a more stable and structured spectral pattern. These visual differences reflect common speech impairments in PD, such as tremor, reduced prosodic variation, and articulatory imprecision.
This analysis supports the role of MFCCs as discriminative features for PD detection and illustrates how neurodegenerative effects are manifested in the spectral-temporal structure of speech.

\paragraph{Intensity}

Intensity refers to the loudness or energy level of a speech signal and is a key prosodic feature in voice analysis. It reflects vocal effort and emotional expressiveness, both of which are often impaired in individuals with PD. The intensity contour, typically computed using root mean square (RMS) energy, tracks how loudness varies over time, with peaks indicating louder segments and valleys representing softer ones.

\begin{figure}[!ht]
    \centering
    \begin{subfigure}[b]{0.53\linewidth} 
        \centering
        \includegraphics[width=\textwidth]{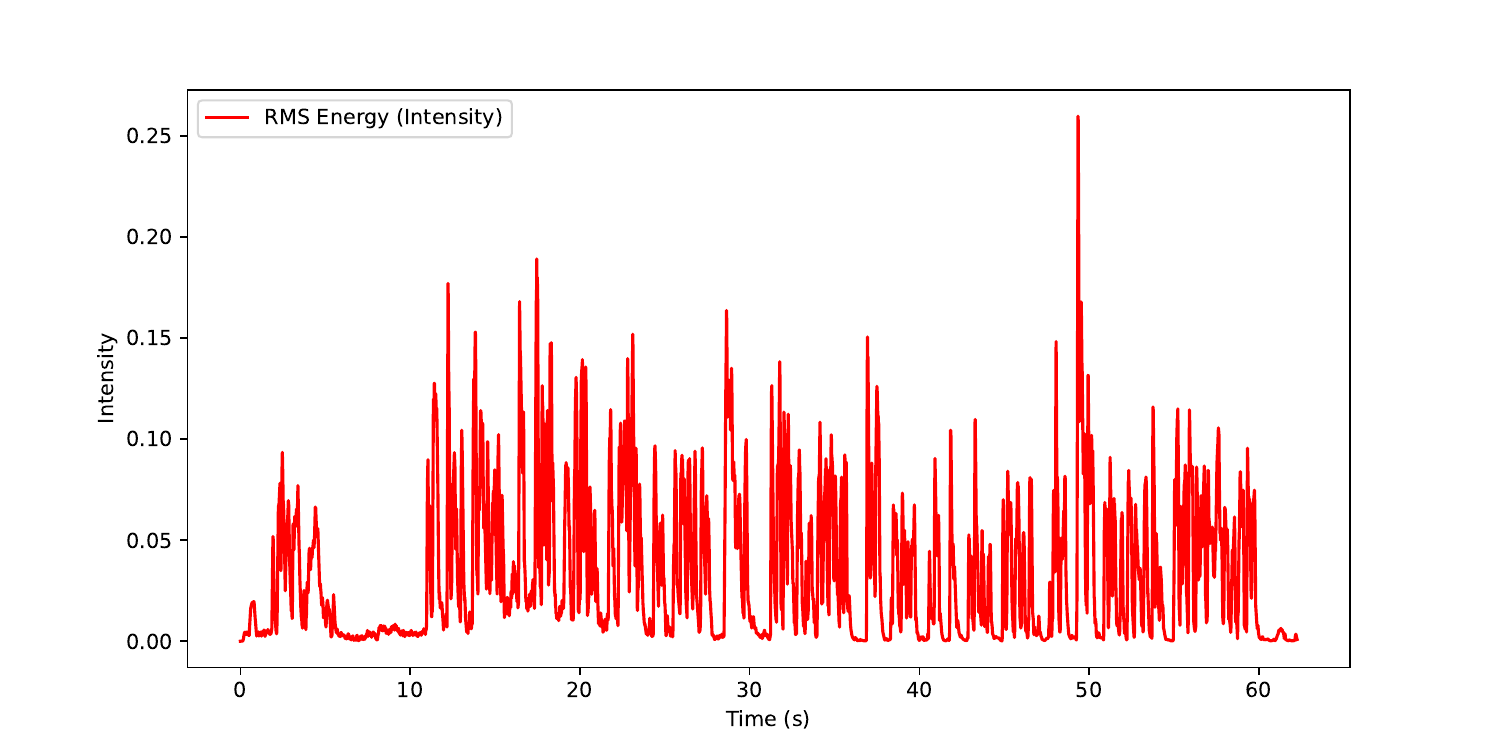}
        \caption{Intensity for a PD.}
        \label{fig:intensity-pd}
    \end{subfigure}
    \begin{subfigure}[b]{0.46\linewidth}
        \centering
        \includegraphics[width=\textwidth]{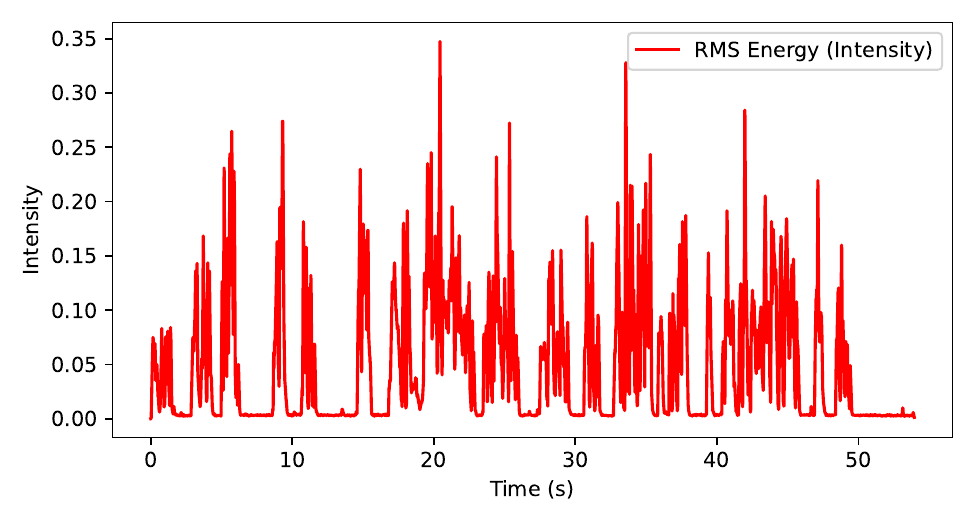}
        \caption{Intensity for an HC.}
        \label{fig:intensity-hc}
    \end{subfigure}
    
    \caption{Intensity profile for a PD and an HC sample. PD and HC correspond to randomly selected subjects from the dataset. The plots are derived from the original raw audio recordings, which exhibit varying durations; therefore, the time axes are not standardized across samples.}
    \label{fig:intensity}
\end{figure}

Figure~\ref{fig:intensity} illustrates the differences in intensity patterns between a PD patient and an HC. The PD curve (Figure~\ref{fig:intensity-pd}) shows reduced and more erratic energy, while the HC curve (Figure~\ref{fig:intensity-hc}) reflects a more controlled and expressive vocal pattern. This contrast reinforces the diagnostic potential of intensity-based features in distinguishing PD-affected speech from healthy speech.

To capture both the absolute and dynamic properties of vocal intensity, we extracted a range of statistical and derivative-based features. These included \texttt{Minimum Intensity}, \texttt{Maximum Intensity}, and \texttt{Mean Intensity}, which provide basic amplitude statistics. We also computed \texttt{Mean Log Energy}, a perceptually motivated transformation of the signal's energy. To model temporal variation and signal dynamics, we further extracted the first-, second-, and third-order derivatives of \texttt{Log Energy}, denoted as $\Delta$\texttt{Log Energy}, $\Delta^2$\texttt{Log Energy}, and $\Delta^3$\texttt{Log Energy}, respectively. These derivative features capture changes in loudness over time and help characterize instability in vocal energy patterns.
\paragraph{Pitch}
Pitch, or the fundamental frequency (F0), is a core prosodic feature in speech that reflects the rate of vocal fold vibration. It plays a crucial role in conveying intonation, emotion, and speaker identity. Variations in pitch, such as its range, stability, and dynamics, can be indicative of both normal linguistic behavior and underlying pathologies. In neurological conditions like PD, pitch irregularities often emerge due to impaired control over the laryngeal musculature, leading to phenomena such as monopitch, pitch breaks, or abnormal pitch range.

\begin{figure}[!ht]
    \centering
    \begin{subfigure}[b]{0.49\linewidth}
    \centering
    \includegraphics[width=\textwidth]{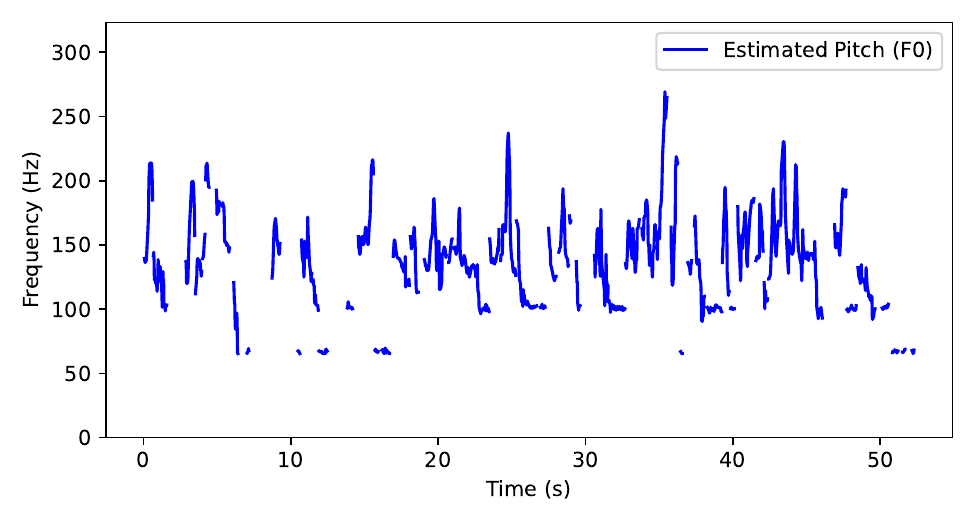}
        \caption{Pitch for a PD}\label{fig:pitch-pd}
    \end{subfigure}
    \begin{subfigure}[b]{0.49\linewidth}
    \centering
    \includegraphics[width=\textwidth]{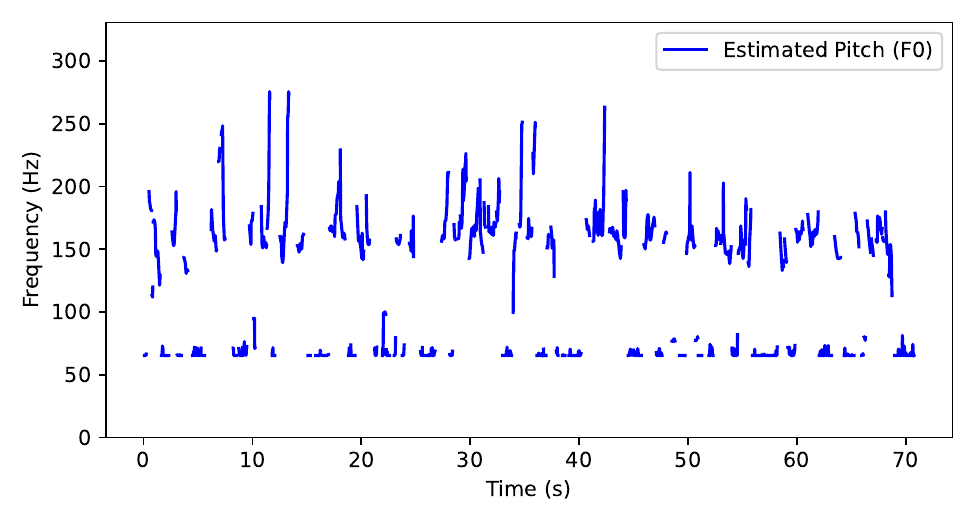}
        \caption{Pitch for an HC}\label{fig:pitch-hc}
    \end{subfigure}
    \caption{Pitch contour for a PD and an HC sample. PD and HC correspond to randomly selected subjects from the dataset. The plots are derived from the original raw audio recordings, which exhibit varying durations; therefore, the time axes are not standardized across samples.}
    \label{fig:pitch}
\end{figure}

Figure~\ref{fig:pitch} compares the estimated pitch contours of a PD patient and an HC. The PD sample (Figure~\ref{fig:pitch-pd}) shows a restricted pitch range with frequent dropouts and irregularities, indicating reduced vocal control and phonatory instability. In contrast, the HC sample (Figure~\ref{fig:pitch-hc}) exhibits a more continuous and dynamic pitch trajectory, reflecting stable voicing and natural prosodic variation. These differences highlight the potential of pitch features, particularly minimum and maximum pitch values, as acoustic markers for distinguishing pathological speech in PD from typical speech patterns.

In this study, we analyze pitch behavior using both \texttt{minimum pitch} and \texttt{maximum pitch} values extracted from the pitch contour. These features provide insight into the individual's vocal range and the ability to modulate pitch effectively. A reduced pitch range or erratic fluctuations can signal reduced prosodic control, commonly observed in PD speech.

\paragraph{Jitter}
Jitter refers to the cycle-to-cycle variation in the fundamental frequency (\texttt{F0}) of a voice signal and serves as a key indicator of pitch instability. It captures the micro-irregularities in the timing of consecutive vocal fold vibrations, often associated with neuromuscular control deficits in the larynx. Elevated jitter levels are commonly observed in individuals with PD, reflecting irregular phonation and reduced vocal fold coordination.

\begin{figure}[!ht]
    \centering
    \begin{subfigure}[b]{0.49\linewidth}
    \centering
    \includegraphics[width=\textwidth]{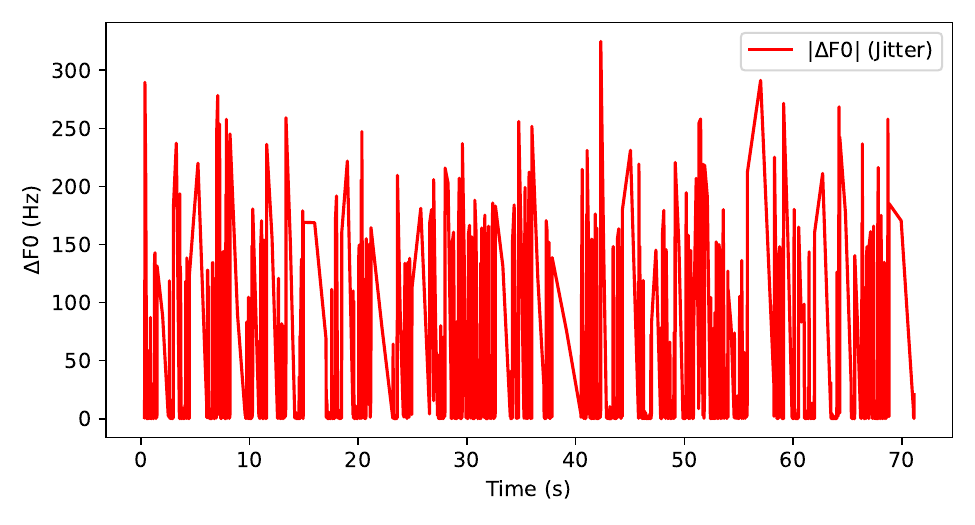}
        \caption{Jitter for a PD}\label{fig:jitter-pd}
    \end{subfigure}
    \begin{subfigure}[b]{0.49\linewidth}
    \centering
    \includegraphics[width=\textwidth]{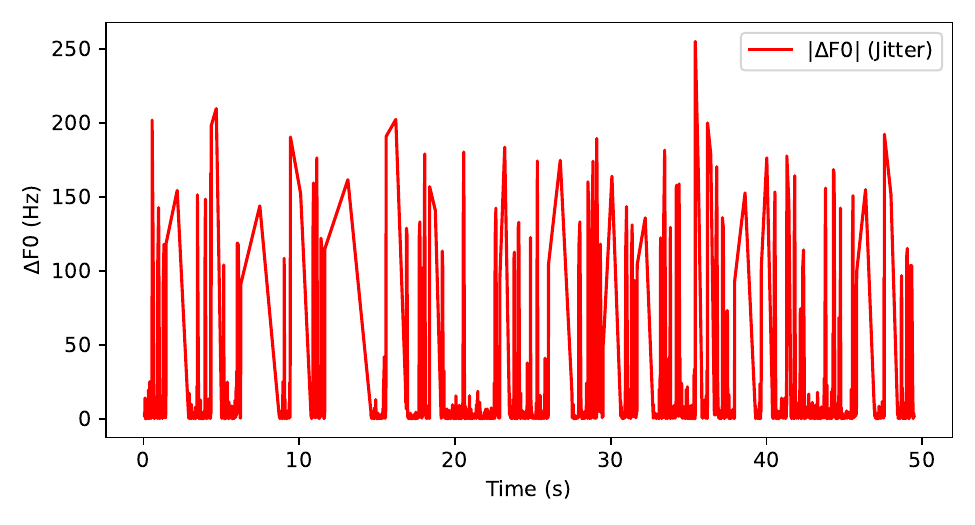}
        \caption{Jitter for an HC}\label{fig:jitter-hc}
    \end{subfigure}
    \caption{Jitter contour for a PD and an HC sample. PD and HC correspond to randomly selected subjects from the dataset. The plots are derived from the original raw audio recordings, which exhibit varying durations; therefore, the time axes are not standardized across samples.}
    \label{fig:jitter}
\end{figure}

Figure~\ref{fig:jitter} presents the jitter contour, representing cycle-to-cycle frequency variation ($\Delta$\texttt{F0}), for a PD patient and an HC. The PD sample (Figure~\ref{fig:jitter-pd}) exhibits frequent and large fluctuations in jitter values, with $\Delta$\texttt{F0} reaching up to 300 Hz. This indicates a high degree of pitch instability and irregular vocal fold vibration, which are characteristic of dysphonia in Parkinsonian speech. The HC sample (Figure~\ref{fig:jitter-hc}) shows a more structured and stable pattern, with lower amplitude jitter values and fewer extreme peaks. While there are occasional fluctuations, they are generally more contained and suggest greater vocal control and smoother pitch modulation.

In this study, we extracted five jitter-related features to comprehensively assess frequency perturbations in speech. These include: (i) Jitter (\texttt{Local}) -- the relative average difference between consecutive pitch periods, expressed as a percentage; (ii) Jitter (\texttt{Local Absolute}) -- the same measure as local jitter but expressed in seconds, providing a time-based representation; (iii) \texttt{RAP} (Relative Average Perturbation) -- the average absolute difference between a period and the mean of its two neighboring periods, offering a smoothed estimate of frequency variability; (iv)
\texttt{PPQ5} (Five-Point Period Perturbation Quotient) -- similar to \texttt{RAP} but calculated over five consecutive periods, capturing mid-term stability; and (v) \texttt{DDP} (Difference of Differences of Periods) -- a derivative metric that amplifies rapid pitch variations, making it sensitive to short-term fluctuations.
These features collectively characterize the degree and nature of pitch instability in speech. Higher values in these jitter metrics are typically associated with dysphonia and aperiodic voice signals, both of which are prevalent in Parkinsonian speech. Incorporating multiple jitter measures enables more nuanced detection of subtle temporal deviations that may not be captured by a single metric alone.

\paragraph{Shimmer}
Shimmer refers to the cycle-to-cycle variation in the amplitude of vocal fold vibrations and serves as a critical marker of amplitude perturbation in voice signals. It quantifies the degree of instability in vocal intensity or loudness over time. Elevated shimmer values are often associated with rough, breathy, or hoarse voice quality, which are common perceptual symptoms of dysphonia in PD. These irregularities typically reflect impaired neuromuscular control over subglottal pressure and vocal fold closure.

\begin{figure}[!ht]
    \centering
    \begin{subfigure}[b]{0.49\linewidth}
    \centering
    \includegraphics[width=\textwidth]{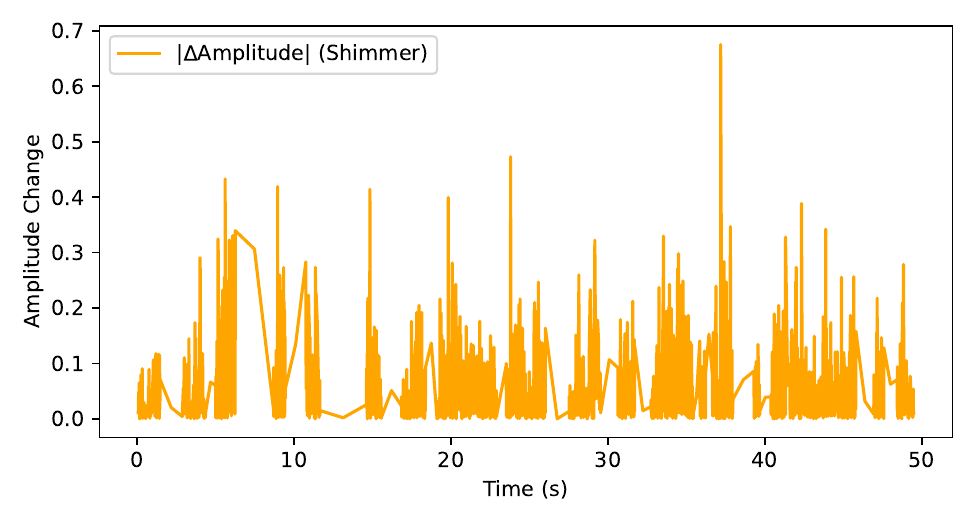}
        \caption{Shimmer for a PD}\label{fig:Shimmer-pd}
    \end{subfigure}
    \begin{subfigure}[b]{0.49\linewidth}
    \centering
    \includegraphics[width=\textwidth]{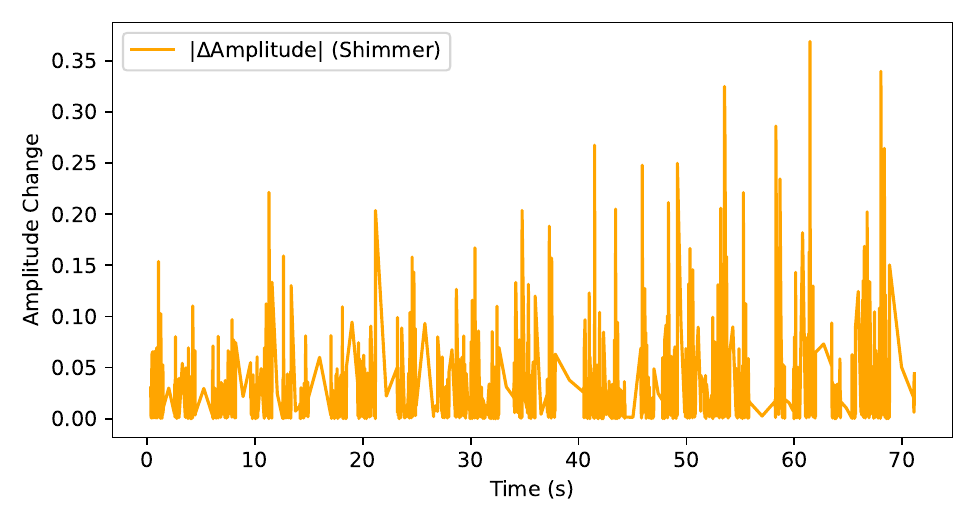}
        \caption{Shimmer for an HC}\label{fig:Shimmer-hc}
    \end{subfigure}
    \caption{Shimmer contour for a PD and an HC sample. PD and HC correspond to randomly selected subjects from the dataset. The plots are derived from the original raw audio recordings, which exhibit varying durations; therefore, the time axes are not standardized across samples.}
    \label{fig:Shimmer}
\end{figure}

Figure~\ref{fig:Shimmer} presents the shimmer contour, measuring cycle-to-cycle amplitude variation, for a PD patient and an HC. In the PD sample (Figure~\ref{fig:Shimmer-pd} the shimmer profile displays frequent and high-magnitude fluctuations in amplitude (reaching up to 0.7) change over time, indicating irregular vocal intensity and unstable phonation, both typical of Parkinsonian speech. In contrast, the HC sample (Figure~\ref{fig:Shimmer-pd}) demonstrates a more stable and consistent shimmer pattern, with lower peak values (mostly below 0.35) and less frequent variation, reflecting greater amplitude control and healthy voice production. This comparison highlights the diagnostic potential of shimmer-based features in identifying speech impairments linked to PD.

To capture a comprehensive picture of amplitude variability, we extracted six shimmer-related features. These were: (i) Shimmer (\texttt{Local}) -- the average absolute difference in amplitude between consecutive cycles, expressed as a percentage; (ii) Shimmer (\texttt{Local dB}) -- the same measure as local shimmer but represented in decibels, offering a logarithmic view of amplitude fluctuations; (iii) \texttt{APQ3} (Amplitude Perturbation Quotient over 3 periods) -- the average absolute amplitude difference between a cycle and the average of its two neighbors; (iv)
\texttt{APQ5} and (vi) \texttt{APQ11} -- similar to APQ3 but computed over 5 and 11 cycles, respectively, enabling mid- and long-term assessment of amplitude perturbations; (vi) \texttt{DDA} (Difference of Differences of Amplitude) -- a derivative feature that emphasizes rapid fluctuations in amplitude across consecutive periods, making it highly sensitive to short-term instability.
These features collectively allow for multi-scale analysis of vocal amplitude irregularities, improving the model's ability to detect subtle voice impairments. Their inclusion is particularly relevant for distinguishing pathological speech patterns in PD, where shimmer-based disruptions often precede other perceptible symptoms.

\paragraph{Formants}
Formant analysis in voice refers to the study of resonant frequencies or spectral peaks in the speech signal, which correspond to the natural resonances of the vocal tract~\citep{SERRURIER2024101374}. These formants are critical for both speech production and perception, as they shape the acoustic properties of different vowels and consonants. PD patient consistently exhibited lower formant frequencies across different syllables \citep{wang2022distinctive}. In this study, we extracted the first four formants (\texttt{f1} to \texttt{f4}) from each speech segment to characterize articulatory behavior.
The first formant (\texttt{f1}) represents the lowest frequency resonance of the vocal tract and is primarily influenced by the size and configuration of the pharyngeal cavity. The second formant (\texttt{f2}) is shaped by the horizontal position of the tongue and the length of the oral cavity. The third formant (\texttt{f3}) contributes to fine-grained vowel distinctions and reflects complex tongue and lip movements. The fourth formant (\texttt{f4}), while less perceptually salient, is associated with overall vocal tract length and contributes to speaker-specific voice quality.

\begin{figure}[!ht]
    \centering
    \begin{subfigure}[b]{0.49\linewidth} 
        \centering
        \includegraphics[width=\textwidth]{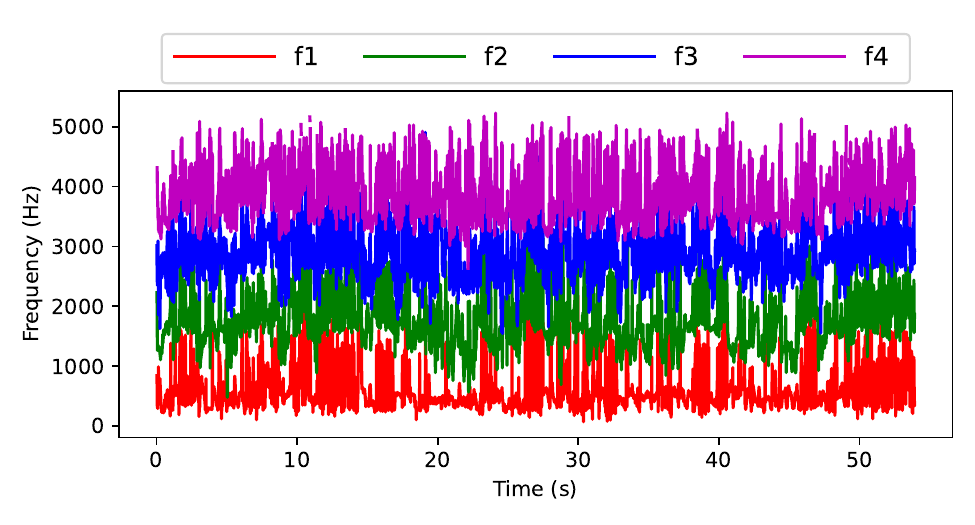}
        \caption{Formants for a PD.}
        \label{fig:formants-pd}
    \end{subfigure}
    \begin{subfigure}[b]{0.49\linewidth}
        \centering
        \includegraphics[width=\textwidth]{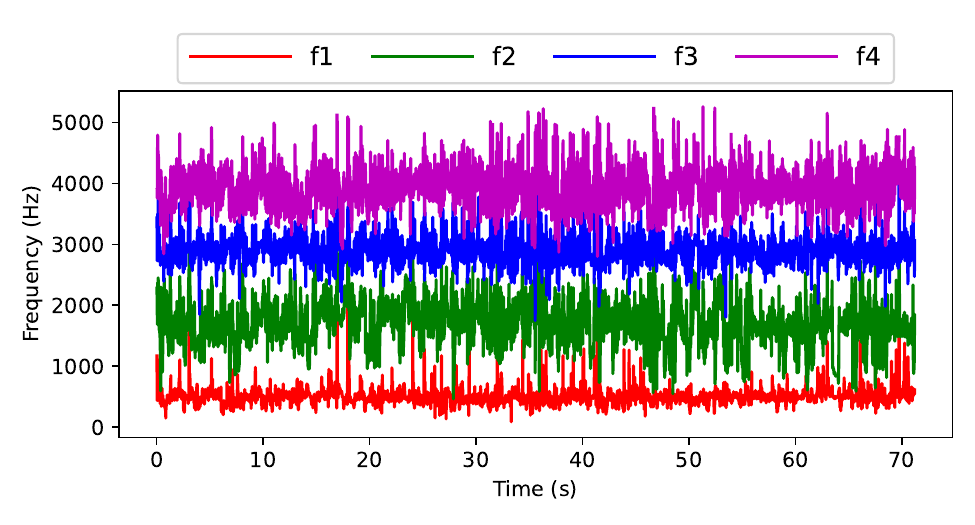}
        \caption{Formants for an HC.}
        \label{fig:formant-hc}
    \end{subfigure}
    
    \caption{Comparison of formant frequency trajectories (f1--f4) for a PD and an HC. PD and HC correspond to randomly selected subjects from the dataset. The plots are derived from the original raw audio recordings, which exhibit varying durations; therefore, the time axes are not standardized across samples.}
    \label{fig:formant}
\end{figure}

Figure~\ref{fig:formant} illustrates the formant frequency profiles (\texttt{f1} to \texttt{f4}) extracted from the speech of a PD patient and an HC. While both profiles exhibit the expected general ordering of formants (\texttt{f1} $<$ \texttt{f2} $<$ \texttt{f3} $<$ \texttt{f4}), the PD sample (Figure~\ref{fig:formants-pd}) shows greater variability and instability, particularly in higher-order formants (\texttt{f3} and \texttt{f4}). These irregularities suggest reduced articulatory precision, a known characteristic of hypokinetic dysarthria in PD.
In contrast, the HC sample (Figure~\ref{fig:formant-hc}) exhibits smoother and more consistent formant trajectories, indicating consistent vocal tract shaping during speech production. This comparison highlights the diagnostic relevance of formant-based features in capturing articulatory deficits in PD speech and supports their inclusion in comprehensive acoustic models for early disease detection.

\paragraph{Pulse}
Analyzing the pulse feature of voice recordings provides valuable insight into the fundamental frequency and temporal characteristics of the recorded speech. It helps assess aspects such as pitch stability, vocal regularity, and overall voice quality, making it essential for voice analysis in various applications, including diagnosing voice disorders such as PD, evaluating vocal function, and monitoring changes in vocal performance over time.

\begin{figure}[!ht]
    \centering
    \begin{subfigure}[b]{0.49\linewidth}
    \centering
    \includegraphics[width=\textwidth]{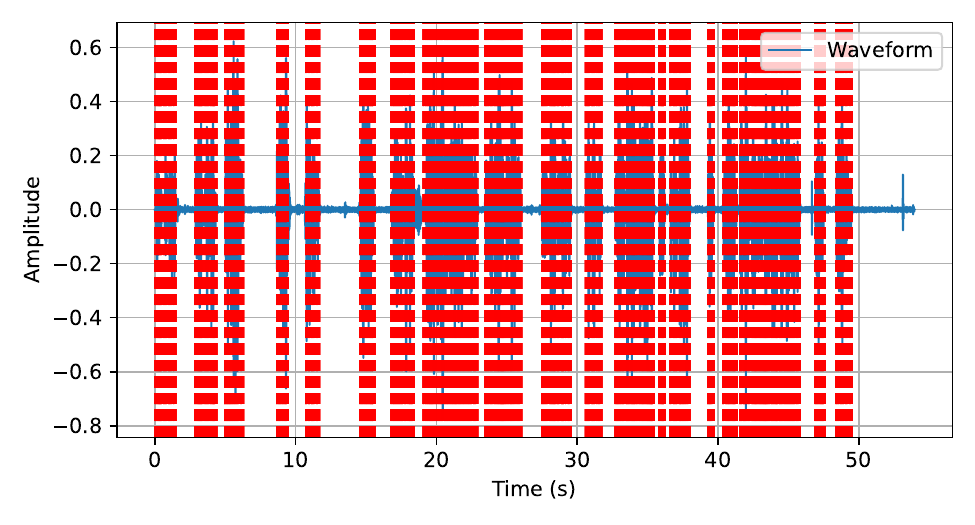}
        \caption{Pulse for a PD}\label{fig:pulse-pd}
    \end{subfigure}
    \begin{subfigure}[b]{0.49\linewidth}
    \centering
    \includegraphics[width=\textwidth]{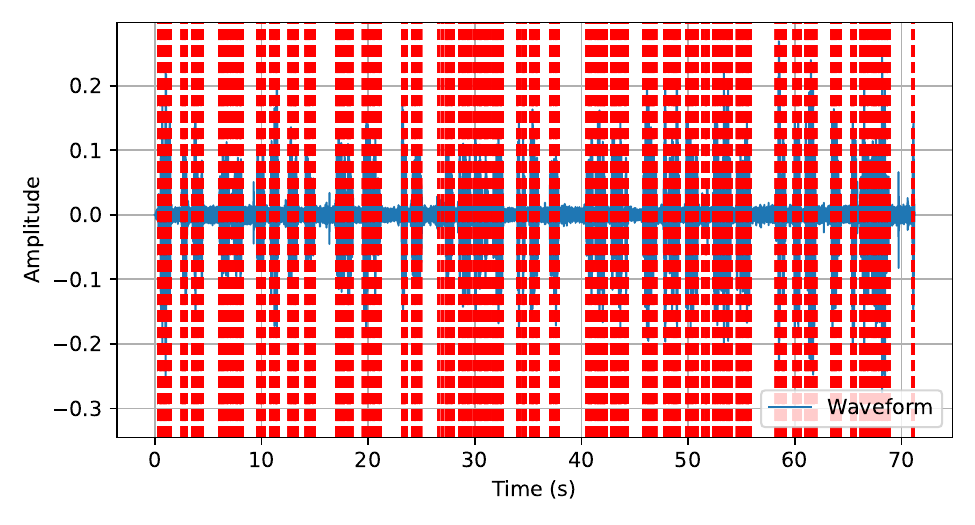}
        \caption{Pulse for an HC}\label{fig:pulse-hc}
    \end{subfigure}
    \caption{Pulse for a PD and an HC sample. PD and HC correspond to randomly selected subjects from the dataset. The plots are derived from the original raw audio recordings, which exhibit varying durations; therefore, the time axes are not standardized across samples.}
    \label{fig:pulse}
\end{figure}

Figure~\ref{fig:pulse} illustrates glottal pulse extraction for a PD patient and an HC, where the waveform is shown in blue and detected glottal pulses are marked with vertical red lines. The PD sample (Figure~\ref{fig:pulse-pd}) exhibits noticeably irregular pulse spacing, suggesting increased variability in vocal fold vibration -- a hallmark of hypophonia and voice instability commonly associated with PD. In contrast, the HC sample (Figure~\ref{fig:pulse-hc}) shows more uniformly spaced pulses, indicative of more stable and periodic vocal fold activity.

In this study, we extracted glottal pulses from the recorded audio and computed key statistical features: the total \texttt{number of pulses}, the \texttt{number of periods} (i.e., intervals between successive pulses), the \texttt{mean period} duration, and the \texttt{standard deviation} of period durations. These measures reflect the degree of temporal regularity in vocal fold vibration.

\paragraph{Harmonicity}
Harmonicity refers to the degree to which a vocal signal comprises harmonically related frequencies, reflecting the regularity and stability of vocal fold vibrations. In voiced speech, periodic vibration generates a fundamental frequency (perceived as pitch) and its harmonics. A high harmonicity level indicates stable phonation with rich harmonic content, while low harmonicity suggests irregular vibrations, breathiness, or increased noise components -- features often associated with disordered speech. \citet{yang2020physical} observed that patients PD exhibited low HNR compared to healthy controls.
In this study, we extracted several harmonicity-related features using autocorrelation-based analysis, including \texttt{AutoCorrHarmonicity} (frame-level harmonicity), \texttt{meanAutoCorrHarmonicity} (signal-level average), \texttt{HNR} (Harmonic-to-Noise Ratio), and its inverse, the \texttt{NHR} (Noise-to-Harmonic Ratio). These measures quantify the relative balance between periodic (harmonic) and aperiodic (noise) components in the voice signal.

\begin{figure}[!ht]
    \centering
    \begin{subfigure}[b]{0.49\linewidth}
    \centering
    \includegraphics[width=\textwidth]{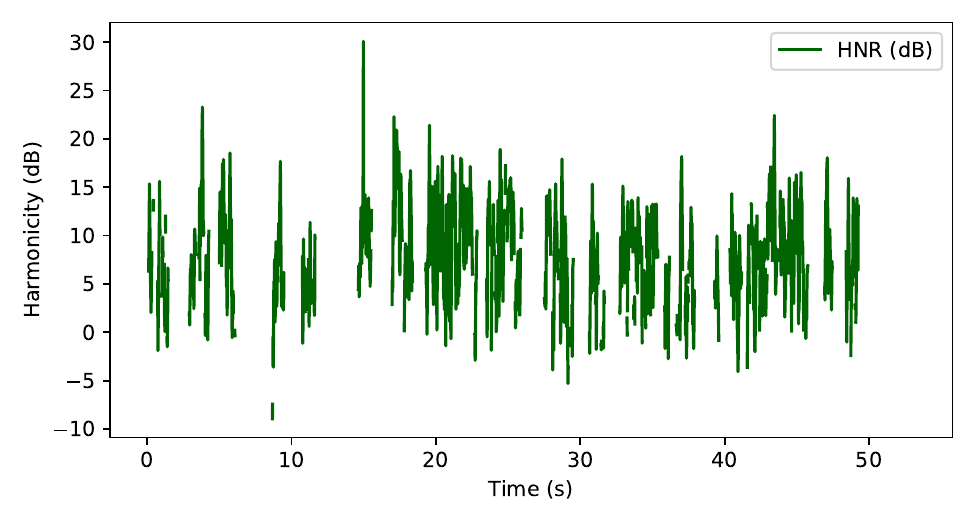}
        \caption{Harmonicity for a PD}\label{fig:harmonicity-pd}
    \end{subfigure}
    \begin{subfigure}[b]{0.49\linewidth}
    \centering
    \includegraphics[width=\textwidth]{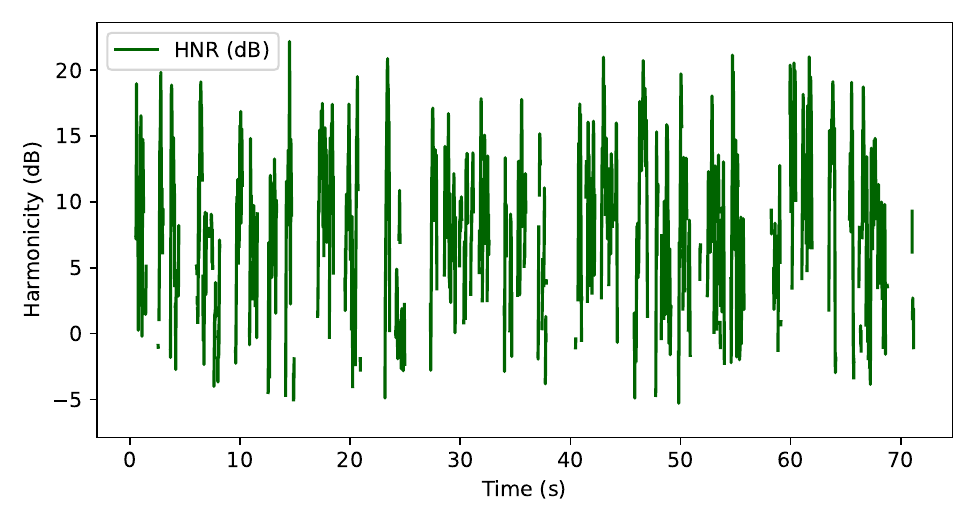}
        \caption{Harmonicity for an HC}\label{fig:harmonicity-hc}
    \end{subfigure}
    \caption{Harmonicity for a PD and an HC sample. PD and HC correspond to randomly selected subjects from the dataset. The plots are derived from the original raw audio recordings, which exhibit varying durations; therefore, the time axes are not standardized across samples.}
    \label{fig:harmonicity}
\end{figure}

Figure~\ref{fig:harmonicity} illustrates the HNR contours for a PD and an HC subject. In the PD sample (Figure~\ref{fig:harmonicity-pd}), the harmonicity curve exhibits frequent and abrupt dips, reflecting high aperiodicity and unstable phonation—hallmarks of Parkinsonian speech. In contrast, the HC sample (Figure~\ref{fig:harmonicity-hc}) displays a more stable HNR trajectory, with fewer fluctuations and consistently higher values, indicating more regular and harmonically rich vocal fold activity typical of healthy speech.

\section{Feature ranking and selection}


To ensure optimal model performance and interpretability, the BenSParX framework incorporates a robust and multi-faceted feature selection strategy. Given the high dimensionality and potential redundancy of acoustic features, effective selection is essential for identifying the most discriminative attributes while minimizing noise and overfitting. BenSParX addresses this by integrating both statistical and algorithmic approaches to feature selection. Specifically, it combines four complementary techniques -- Recursive Feature Elimination with Cross-Validation (RFECV), Least Absolute Shrinkage and Selection Operator (LASSO), Relief-F, and the Mann-Whitney U test -- each offering unique strengths in evaluating feature relevance. This integrated approach enables the framework to construct a lean, high-impact feature set tailored for PD detection.

\subsection{Recursive Feature Elimination with Cross-Validation (RFECV)}
\begin{figure}[!ht]
    \centering
    \includegraphics[width=.9\textwidth]{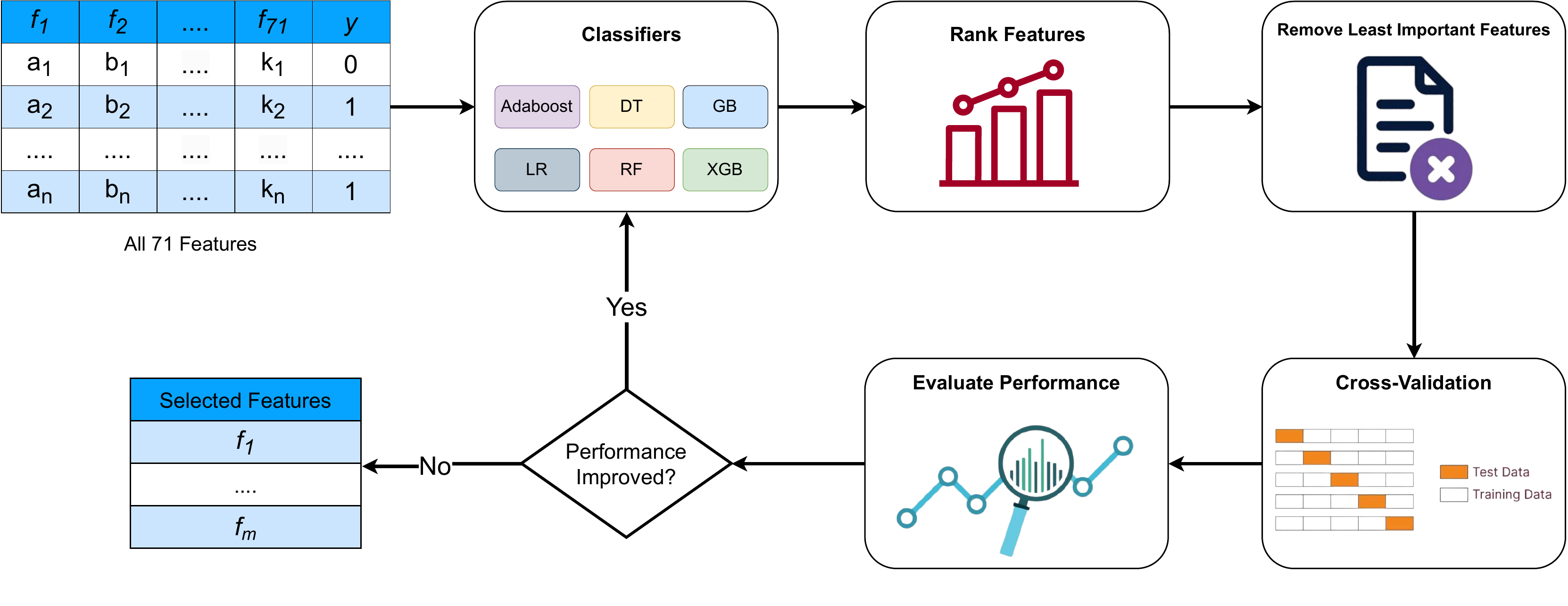} 
    \caption{An outline of the RFECV procedure for feature selection using Classifier.}
    \label{fig:rfecv}
\end{figure}

RFECV is a wrapper-based feature selection technique that recursively eliminates the least important features based on model performance until an optimal subset is identified. It extends the traditional recursive feature elimination (RFE) approach by incorporating k-fold cross-validation to automatically determine the number of features that maximize performance, as illustrated in Figure \ref{fig:rfecv}. RFECV uses underlying machine learning classifiers to rank features based on their importance scores. At each iteration, the model evaluates performance using the current subset of features and assigns importance rankings (e.g., via feature weights or impurity-based scores). The least important feature(s) are eliminated, and the model is re-trained and re-evaluated. This recursive elimination process continues until the optimal number of features is identified, as determined by cross-validation performance. By coupling feature ranking with iterative pruning and validation, RFECV ensures that only the most relevant features are retained for the final model. In the BenSParX framework, we employed six diverse classifiers within the RFECV process, AdaBoost, Decision Tree, Gradient Boosting, Logistic Regression, Random Forest, and XGBoost, to conduct a comprehensive and classifier-agnostic feature selection.

At each iteration $t$, RFECV evaluates a candidate feature subset $S_t$ using a scoring function based on the cross-validation loss:
\begin{equation}
E(S_t) = \frac{1}{k} \sum_{i=1}^{k} \mathcal{L}(y_i, f_{\theta}(X_i))
\end{equation}
The process continues until the optimal subset $S^*$ is found:
\begin{equation}
S^* = \arg\min_{S_t} E(S_t)
\end{equation}

In high-dimensional datasets, where redundant or unnecessary features could impair model performance, RFECV is especially helpful. By integrating cross-validation, RFECV ensures that the selected features generalize well to unseen data, making it a robust choice for feature selection in machine learning pipelines like BenSParX.

\subsection{Least Absolute Shrinkage and Selection Operator (LASSO)}
LASSO is a regularization-based feature selection method that enhances model performance by introducing sparsity in the feature space. LASSO is particularly useful for high-dimensional datasets, where the number of features may exceed or approach the number of observations. By applying an $\ell_1$ penalty, LASSO automatically shrinks the coefficients of less informative features to exactly zero, effectively removing them from the model and yielding a compact, interpretable subset of predictors.
The LASSO optimization problem is defined as:
\begin{equation}
\hat{\beta} = \arg\min_{\beta} \left( \sum_{i=1}^{n} \left( y_i - \sum_{j=1}^{p} X_{ij} \beta_j \right)^2 + \lambda \sum_{j=1}^{p} |\beta_j| \right)
\end{equation}
Here, the $\ell_1$ penalty term $\lambda \sum_{j=1}^{p} |\beta_j|$ forces some coefficients $\beta_j$ to become exactly zero, effectively removing less relevant features from the model. As $\lambda$ increases, more features are eliminated, leading to a sparse model that retains only the most significant predictors. The optimal value of $\lambda$ is typically determined using cross-validation:

\begin{equation}
\lambda^* = \arg\min_{\lambda} \sum_{i=1}^{k} \mathcal{L}(y_i, f_{\theta}(X_i))
\end{equation}

LASSO is widely used in feature selection due to its ability to handle collinearity and improve model interpretability by selecting only the most relevant features. This makes it a valuable technique in machine learning frameworks, such as BenSParX, for isolating the most discriminative voice features for PD detection.

\subsection{Relief-F with Sequential Forward Selection}
\begin{figure}[!ht]
    \centering
    \includegraphics[width=.9\textwidth]{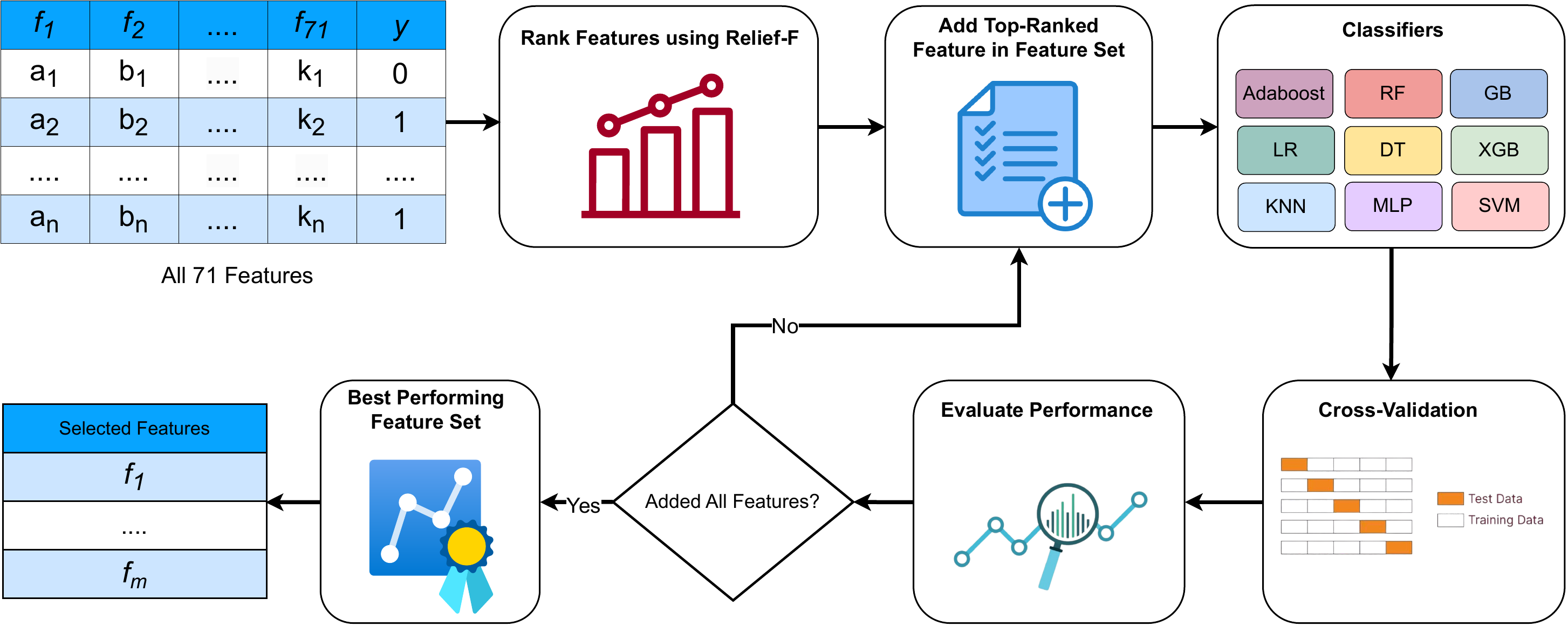} 
    \caption{An outline of the Relief-F procedure for feature selection using Classifier.}
    \label{fig:relieff}
\end{figure}

We employed a hybrid selection strategy that combines ReliefF-based ranking with Sequential Forward Selection (SFS) guided by cross-validation, as illustrated in Figure~\ref{fig:relieff}. ReliefF was first applied to the entire feature space to compute importance scores based on each feature's ability to distinguish between instances of different classes, in our case, PD and HC. This produced a ranked list of features, prioritizing those most relevant to PD classification.

Relief-F is a filter-based algorithm that estimates feature importance based on how well features differentiate between neighboring instances of different classes. It is a more sophisticated feature selection technique that builds upon the Relief algorithm by handling multi-class problems and noisy datasets. The algorithm iteratively selects a random instance and finds its **nearest neighbors**, including:
\begin{itemize}
    \item The nearest **hit** ($H$) – a sample from the same class.
    \item The nearest **misses** ($M_c$) – samples from different classes.
\end{itemize}
The weight of each feature $f_j$ is updated using the following equation:
\begin{equation}
W(f_j) = W(f_j) - \sum_{c \neq C} P(c) \cdot \left( \frac{ \text{diff}(f_j, X, H)}{m} - \frac{ \text{diff}(f_j, X, M_c)}{m} \right)
\end{equation}
The difference function $\text{diff}(f_j, X, Y)$ is defined as:

\begin{equation}
\text{diff}(f_j, X, Y) =
\begin{cases}
|X_j - Y_j|, & \text{if } f_j \text{ is numerical} \\
0, & \text{if } X_j = Y_j \text{ (categorical)} \\
1, & \text{otherwise}
\end{cases}
\end{equation}

After iterating through multiple instances, features with higher $W(f_j)$ values are considered more important, while those with lower values contribute less to classification. Building on this ranking, SFS was then used to iteratively construct the optimal subset of features. SFS is a wrapper-based approach that requires a classifier to evaluate model performance during the selection process. In the BenSParX framework, we integrated nine diverse classifiers -- AdaBoost, Decision Tree, Gradient Boosting, K-Nearest Neighbors (KNN), Logistic Regression, Multilayer Perceptron (MLP), Random Forest, Support Vector Machine (SVM), and XGBoost -- to comprehensively assess each feature's contribution to classification accuracy. At each iteration, the feature whose inclusion resulted in the greatest improvement in cross-validation performance was retained. This process continued until the maximum subset size was reached, after which the best-performing feature subset was identified based on the highest overall classification performance. 
By combining the efficiency of filter-based ranking with the classification-driven rigor of SFS across multiple learners, this hybrid approach ensured the selection of a robust, high-performing, and generalizable feature subset for PD detection.


\subsection{Mann-Whitney U Test}
The Mann-Whitney U test is a non-parametric statistical method used to assess whether two independent groups are drawn from the same distribution. In feature selection, it is particularly useful for identifying features that exhibit statistically significant differences between two classes, such as the PD and HC groups. The \textit{U} statistic is calculated based on the ranks of feature values from both groups. For class 1, the statistic is computed as:
\begin{equation}
U = n_1 n_2 + \frac{n_1 (n_1 + 1)}{2} - R_1
\end{equation}
Similarly, the second U statistic for class 2 is computed as:
\begin{equation}
U = n_1 n_2 + \frac{n_2 (n_2 + 1)}{2} - R_2
\end{equation}
Here, $n_1$ and $n_2$ are sample sizes of class 1 and class 2, and $R_1$ and $R_2$ are the sum of the ranks of class 1 and class 2, respectively. The smaller of the two \textit{U} values is used to determine statistical significance. A \textit{p}-value is derived from the \textit{U} statistic to assess whether the observed difference in feature distributions is significant. A feature is considered important if its \textit{p}-value is below a predefined significance threshold (e.g., $p<0.05$). Features with lower \textit{p}-values exhibit stronger class differentiation and are prioritized for model training. The Mann-Whitney \textit{U} test offers a simple yet effective univariate filtering method, especially in cases where feature distributions do not meet parametric assumptions, making it a valuable component of the BenSParX framework for robust, statistically grounded feature selection.

\section{Feature-rank stability across outer cross-validation folds}
\label{subsec:shap_rank_stability}

Table~\ref{tab:shap_rank_stability} reports, for every feature that was
selected in at least one outer fold: (i) the number of outer folds (out
of 10) in which the Mann-Whitney U test selected it, mirroring the
fold-level counts already reported by category; (ii) its mean SHAP-importance rank,
averaged across the folds in which it was selected; and (iii) the range
(minimum--maximum) of that rank across folds. A narrow rank range
alongside selection in all 10 folds indicates that a feature's
importance is consistent regardless of which participants happened to
fall into the training versus test partition, whereas a wide range or
partial selection frequency indicates that a feature's apparent
importance is comparatively fold-dependent.
 
The six most important features -- \texttt{std\_delta\_delta\_log\_energy},
\texttt{std\_delta\_log\_energy}, \texttt{Delta\_MFCC\_2}, \texttt{MFCC\_2},
\texttt{meanAutoCorrHarmonicity}, and \texttt{meanPeriodPulses} -- were
selected in all 10 outer folds, with mean SHAP rank varying by no more
than a few positions across folds (e.g.,
\texttt{std\_delta\_delta\_log\_energy} ranked first in every fold). This
stability supports the interpretation, that energy-variability,
spectral-envelope (MFCC), and harmonicity features are robust,
fold-independent markers of Parkinsonian speech rather than artifacts of
a particular data split. A small number of features (e.g.,
\texttt{Delta\_MFCC\_1}, \texttt{MFCC\_1}, \texttt{Delta\_MFCC\_11},
\texttt{Delta\_MFCC\_12}, \texttt{MFCC\_11}, \texttt{MFCC\_12}) were
selected in only 3 of the 10 folds, consistent with the fold-dependent
variability already noted for individual MFCC coefficients.
 
\begin{longtable}{l c c c}
\caption{Feature-rank stability across the 10 outer cross-validation
folds. \emph{Folds selected} is the number of outer folds (out of 10)
in which the Mann-Whitney U test selected the feature; \emph{mean rank}
is the feature's mean SHAP-importance rank across those folds;
\emph{rank range} is the minimum--maximum of that rank across folds.
Features are sorted by folds selected (descending), then by mean rank
(ascending).}
\label{tab:shap_rank_stability} \\
\toprule
Feature & Folds selected & Mean rank & Rank range \\
\midrule
\endfirsthead
 
\multicolumn{4}{c}{\tablename\ \thetable\ -- continued from previous page} \\
\toprule
Feature & Folds selected & Mean rank & Rank range \\
\midrule
\endhead
 
\midrule
\multicolumn{4}{r}{\emph{Continued on next page}} \\
\endfoot
 
\bottomrule
\endlastfoot
 
std\_delta\_delta\_log\_energy & 10/10 & 1.0 & 1--1 \\
std\_delta\_log\_energy & 10/10 & 2.4 & 2--3 \\
Delta\_MFCC\_2 & 10/10 & 3.0 & 2--4 \\
MFCC\_2 & 10/10 & 3.7 & 3--5 \\
meanAutoCorrHarmonicity & 10/10 & 5.0 & 4--6 \\
meanPeriodPulses & 10/10 & 6.6 & 5--8 \\
std\_Log\_energy & 10/10 & 8.3 & 6--20 \\
rapJitter & 10/10 & 10.9 & 7--15 \\
Delta\_MFCC\_3 & 10/10 & 10.9 & 7--13 \\
minIntensity & 10/10 & 11.1 & 6--15 \\
MFCC\_3 & 10/10 & 11.1 & 6--17 \\
apq11Shimmer & 10/10 & 11.9 & 8--19 \\
ddpJitter & 10/10 & 13.7 & 7--25 \\
DeltaDelta\_MFCC\_8 & 10/10 & 16.1 & 11--24 \\
numPulses & 10/10 & 16.4 & 9--23 \\
numPeriodsPulses & 10/10 & 16.5 & 9--23 \\
maxPitch & 10/10 & 16.7 & 7--30 \\
DeltaDelta\_MFCC\_7 & 10/10 & 19.1 & 12--28 \\
locPctJitter & 10/10 & 19.3 & 10--26 \\
Delta\_MFCC\_7 & 10/10 & 20.1 & 17--26 \\
MFCC\_7 & 10/10 & 20.5 & 16--29 \\
meanIntensity & 10/10 & 21.3 & 16--26 \\
ppq5Jitter & 10/10 & 25.6 & 12--36 \\
locAbsJitter & 10/10 & 25.6 & 20--32 \\
DeltaDelta\_MFCC\_4 & 10/10 & 25.8 & 21--29 \\
f1 & 10/10 & 25.8 & 19--29 \\
maxIntensity & 10/10 & 29.5 & 21--36 \\
Delta\_MFCC\_4 & 10/10 & 32.7 & 28--38 \\
MFCC\_4 & 10/10 & 33.1 & 25--37 \\
DeltaDelta\_MFCC\_9 & 10/10 & 33.9 & 28--39 \\
MFCC\_8 & 10/10 & 34.0 & 28--40 \\
MFCC\_0 & 10/10 & 34.3 & 28--39 \\
stdDevPeriodPulses & 10/10 & 35.5 & 28--42 \\
Delta\_MFCC\_8 & 10/10 & 35.5 & 28--43 \\
Delta\_MFCC\_0 & 10/10 & 36.1 & 31--43 \\
DeltaDelta\_MFCC\_6 & 10/10 & 38.7 & 33--48 \\
DeltaDelta\_MFCC\_10 & 10/10 & 40.4 & 33--55 \\
f3 & 10/10 & 41.2 & 34--49 \\
DeltaDelta\_MFCC\_3 & 10/10 & 41.2 & 31--48 \\
f2 & 10/10 & 42.4 & 30--51 \\
DeltaDelta\_MFCC\_11 & 10/10 & 43.2 & 35--54 \\
meanNoiseToHarmHarmonicity & 10/10 & 46.3 & 41--55 \\
MFCC\_5 & 10/10 & 47.0 & 43--52 \\
Delta\_MFCC\_5 & 10/10 & 47.4 & 43--56 \\
meanHarmToNoiseHarmonicity & 10/10 & 49.0 & 42--57 \\
DeltaDelta\_MFCC\_5 & 10/10 & 49.5 & 44--60 \\
meanPitch & 10/10 & 50.3 & 36--57 \\
locDbShimmer & 10/10 & 51.2 & 39--60 \\
DeltaDelta\_MFCC\_12 & 10/10 & 53.1 & 46--60 \\
MFCC\_9 & 10/10 & 53.3 & 44--61 \\
mean\_Log\_energy & 10/10 & 54.2 & 40--61 \\
Delta\_MFCC\_9 & 10/10 & 54.2 & 48--58 \\
locShimmer & 10/10 & 55.1 & 49--61 \\
apq5Shimmer & 10/10 & 55.8 & 47--62 \\
DeltaDelta\_MFCC\_1 & 10/10 & 56.4 & 50--60 \\
DeltaDelta\_MFCC\_0 & 10/10 & 58.0 & 46--62 \\
ddaShimmer & 10/10 & 58.5 & 54--63 \\
apq3Shimmer & 10/10 & 58.7 & 53--64 \\
Delta\_MFCC\_6 & 9/10 & 21.1 & 12--27 \\
MFCC\_6 & 9/10 & 22.1 & 15--26 \\
Delta\_MFCC\_1 & 3/10 & 24.7 & 20--30 \\
MFCC\_1 & 3/10 & 25.7 & 21--31 \\
Delta\_MFCC\_12 & 3/10 & 41.7 & 40--43 \\
MFCC\_12 & 3/10 & 44.7 & 42--49 \\
Delta\_MFCC\_11 & 3/10 & 48.7 & 44--53 \\
MFCC\_11 & 3/10 & 54.0 & 51--57 \\
 
\end{longtable}

\bibliographystyle{elsarticle-harv}

\bibliography{reference}

\end{document}